%% file: main.tex
\documentclass{article}
\usepackage{arxiv}
\usepackage[american]{babel}

\usepackage{natbib} 
\usepackage{mathtools} 
\usepackage{booktabs} 
\usepackage{tikz} 
\usepackage{definitions}
\usepackage{thmtools, thm-restate}
\usepackage{hyperref}
\usepackage{url}
\usepackage{float}
\usepackage{pdfpages}
\usepackage{wrapfig}
\usepackage{enumitem}

\newcommand{\method}{\text{OSLow}}
\title{Order-based Structure Learning with Normalizing Flows}

\author{%
  \href{mailto:<hamidrezakamkari@gmail.com>?Subject=Your paper OSLow}{Hamidreza Kamkari}\thanks{Equal Contribution. Our code is accessible at \href{https://github.com/HamidrezaKmK/oslow}{\url{https://github.com/HamidrezaKmK/oslow}}}  \quad Vahid Balazadeh$^\star$ \quad Vahid Zehtab\quad Rahul G. Krishnan\\
  University of Toronto, Vector Institute
}
\date{}

\begin{document}
\maketitle

\begin{abstract}
\input sections/abstract
\end{abstract}

\input sections/introduction
\input sections/problem-setup
\input sections/related
\input sections/method
\input sections/experiments
\input sections/conclusion

\bibliography{main}

\newpage

\appendix
\input appendix

\end{document}

%% file: sections/abstract.tex
Estimating the causal structure of observational data is a challenging combinatorial search problem that scales super-exponentially with graph size. Existing methods use continuous relaxations to make this problem computationally tractable but often restrict the data-generating process to additive noise models (ANMs) through explicit or implicit assumptions. We present \textbf{O}rder-based \textbf{S}tructure \textbf{L}earning with Normalizing Fl\textbf{ow}s (\method), a framework that relaxes these assumptions using autoregressive normalizing flows. We leverage the insight that searching over topological orderings is a natural way to enforce acyclicity in structure discovery and propose a novel, differentiable permutation learning method to find such orderings. Through extensive experiments on synthetic and real-world data, we demonstrate that \method\ outperforms prior baselines and improves performance on the observational Sachs and SynTReN datasets as measured by structural hamming distance and structural intervention distance, highlighting the importance of relaxing the ANM assumption made by existing methods.

%% file: sections/introduction.tex
\section{Introduction}
Identifying the direction of cause and effect is a fundamental challenge in numerous fields such as genetics, economics, and healthcare~\citep{sachs2005causal,pearl2009causality,zhang2013integrated}. A common method for encapsulating cause-effect relationships between multiple components is to define directed acyclic graphs (DAGs). These structures offer practitioners a means to visualize inferred relationships, reduce computational complexity by disregarding irrelevant associations, and formulate testable scientific hypotheses by identifying effective relationships for intervention.
Two central challenges arise for learning DAGs: (i) The combinatorial nature of the super-exponential space of DAGs renders exhaustive search infeasible, and (ii) determining the causal structure from observations alone is an ill-posed problem; identical data distributions can originate from data-generating processes with different structures.

To address the first challenge, many algorithms define a suitable score over the space of DAGs and aim to find a structure that maximizes it. This search problem is NP-hard~\citep{chickering2004large}; therefore, heuristic approaches are developed based on greedy search such as PC, FCI, and GES~\citep{bouckaert1992optimizing,singh1993algorithm,spirtes2000causation,chickering2002optimal,friedman2003being,zhang2008completeness,teyssier2012ordering,buhlmann2014cam,scanagatta2015learning,park2017bayesian}. Being combinatorial in nature, these approaches do not take advantage of gradient-based optimization. Thus, a second class of algorithms emerged expressing structure learning as a continuous optimization to minimize a differentiable loss subject to soft acyclicity constraints (NOTEARS, GraNDAG)~\citep{zheng2018dags,lachapelle2019gradient,zheng2020learning}. Despite their success, they do not have a hard guarantee of acyclicity. Therefore, a third class of order-based methods is proposed by first learning the topological ordering of variables and then constructing the DAG with respect to the learned ordering, thereby naturally preventing the occurrence of cycles in the predicted structure~~\citep{buhlmann2014cam,charpentier2022differentiable,zantedeschi2023dag}. Interestingly, the first step of order-based methods alone can provide multiple benefits. For example, \cite{buhlmann2014cam} demonstrate that just by knowing the causal ordering, one can accurately estimate the interventional distribution, rendering it an appropriate choice for applications such as average treatment effect estimation~\citep{geffner2022deep}.

Identifiability is the second challenge; causal discovery from observational data is impossible without assumptions. Making explicit interventions akin to those conducted in scientific experiments can circumvent this issue but can be impractical, costly, or ethically questionable. Researchers, therefore, typically make assumptions about the data-generating process to ensure a unique causal structure generates the given observational data. A common assumption is that the noise introduced during data generation is additive. With minor stipulations, additive noise models (ANMs) produce data whence the causal structure is identifiable~\citep{peters2014causal}. Although methods for structure identification via continuous optimization do not explicitly highlight this assumption being made, their loss function implicitly enforces it, as we later discuss in the paper. Recently, \citet{khemakhem2021causal,strobl2022identifying,immer2022identifiability} proved the identifiability of causal structures for a more general class of models than ANMs, called location-scale noise models (LSNMs), which considers heteroscedastic noise. However, most proposed structure learning methods for this class of models solely work for bivariate settings. This motivates provably correct, \emph{and} practical learning algorithms that can discover the causal structure of various data-generating processes, including ANMs and LSNMs or even larger classes of models. Moreover, can relaxing the ANM assumption provide quantifiable benefits to structure learning in real-world datasets, where the true model class is often unknown?

\begin{figure*}[t]
    \centering
    \includegraphics[width=0.75\textwidth]{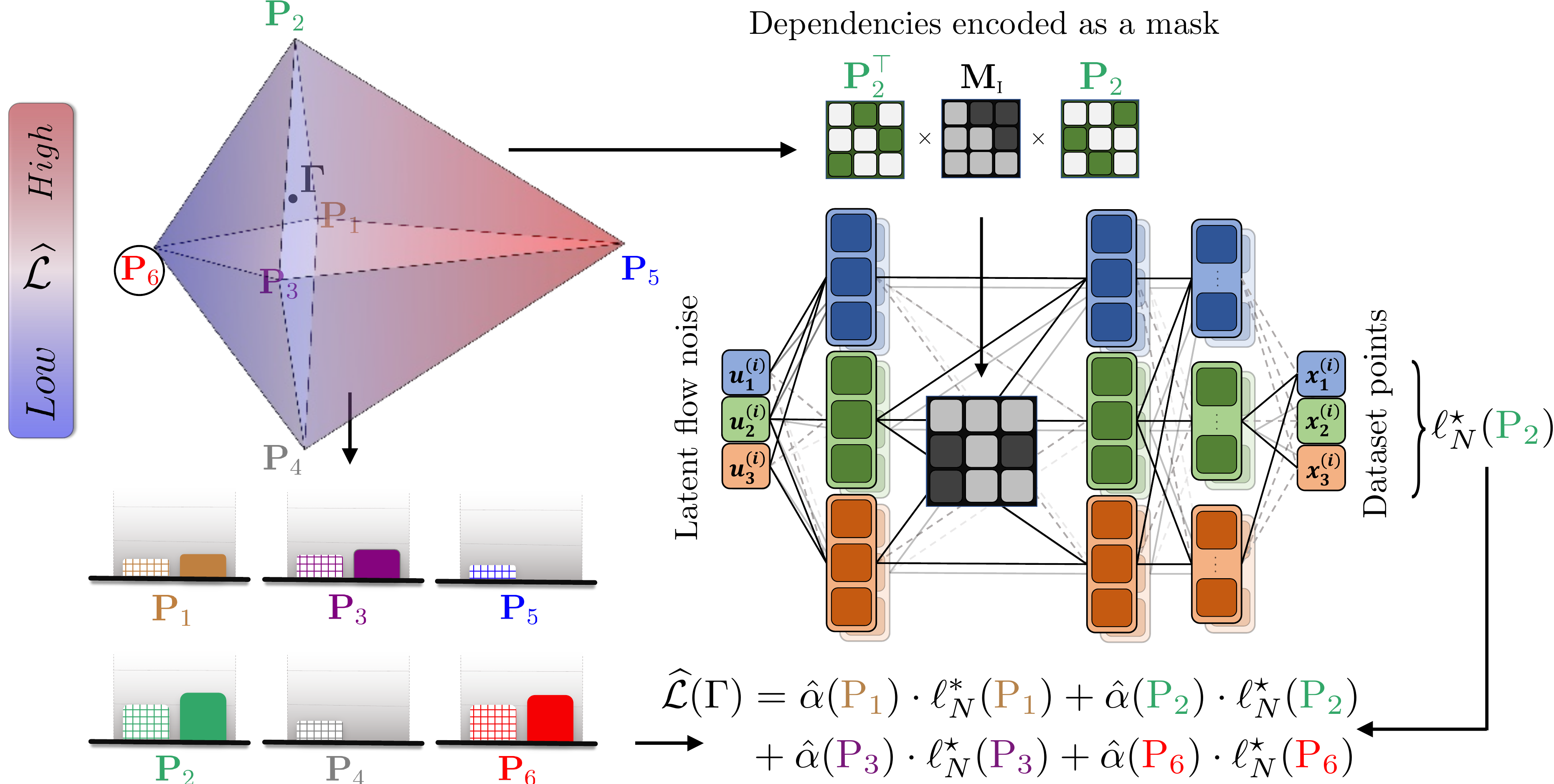}
    \caption{{\small A summary of \method\ for a ($d$ = 3) dimensional structure learning example that is formulated as searching over the set of all possible permutation matrices $\{\vect{P}_i\}_{i=1}^{d!}$.
    \textbf{Upper Right:} Demonstrating how a permutation matrix $\vect{P}_i$ shapes dependencies in our ANF architecture, leading to the formulation of the negative log-likelihood $\ell^*_N(\vect{P}_i)$ corresponding to that ordering on the entire observational dataset of size $N$.
\textbf{Lower Left:} A Boltzmann distribution $\alpha(\cdot)$ is defined over all the permutations and is parameterized by $\vect{\Gamma} \in \R^{d\times d}$ with the energy corresponding to $\vect{P}_i$ defined as the inner product of $\vect{\Gamma}$ and $\vect{P}_i$. We estimate the Boltzmann probability masses (detailed in the method) and set it to zero for permutations that are unlikely to be sampled. Hashed bars indicate true probability mass $\alpha(\cdot)$, while solid bars show the estimated probability $\hat{\alpha}(\cdot)$.
\textbf{Lower Right:} The combination of the estimated distribution and flow likelihoods result in an overall differentiable loss $\widehat{\mathcal{L}}$ for order learning.}
    \textbf{Upper Left:} Visualizing $\widehat{\mathcal{L}}$ over the entire Birkhoff polytope. The blue hue indicates how close are we to an ordering that produces the maximum likelihood with $\vect{P}_6$ being that ordering.
}
    \label{fig:main_figure}
\end{figure*}

We pose the estimation of the correct topological ordering as inferring a permutation over the random variables and introduce \method, a two-fold order-based structure learning method that combines the following ideas. First, we devise an autoregressive normalizing flow (ANF) architecture that simultaneously models the data-generating process for multiple autoregressive orderings by taking the permutation matrix corresponding to that ordering as input. Second, to search over the super-exponential space of topological orderings, we leverage a parameterized Boltzmann distribution over the permutation matrices (i.e., the vertices of a Birkhoff polytope) with energy functions defined as the distance to each vertex. We then define our energy-based loss function as the expected negative log-likelihood of the permutations under this distribution, which provides a differentiable optimization landscape for searching over the space of orderings. In particular, we propose a novel approximation method for estimating this loss that improves over other commonly used approximated permutation learning methods, such as the Gumbel-Sinkhorn method~\citep{mena2018learning}. See \autoref{fig:main_figure} for an illustration.

\xhdr{Contributions}  (i) We design a scalable order-based algorithm based on ANFs that can provably extract valid causal structures from observational data and use it to model interventional distributions; (ii) we introduce a novel permutation learning algorithm that facilitates gradient-based optimization and demonstrate its effectiveness over baselines for causal structure discovery, and (iii) we present improved results on the causal discovery benchmarks Sachs~\citep{sachs2005causal} and SynTReN~\citep{van2006syntren}, thereby emphasizing the importance of relaxing the ANM assumption for structure learning on real data.

%% file: sections/problem-setup.tex
\section{Problem Setup}
\label{sec:problem-set}
\xhdr{Data-Generating Model} We formulate the data distribution using a structural causal model (SCM) $\scm = (\graph, \Fcal, P_{\vect{U}})$, where $\graph$ is the causal graph, $\Fcal$ is a family of $d$ deterministic link functions, and $P_{\vect{U}}$ is a distribution over mutually independent exogenous noise variables $\vect{U} = (U_1, \hdots, U_d)$. $\scm$ entails an observational distribution $P_\vect{X}$ over the random variables $\vect{X} = (X_1, \ldots, X_d)$, where each random component is obtained by evaluating the link function $f_i \in \Fcal$ over the corresponding exogenous noise $U_i \sim P_{\vect{U}_i}$ and the parents of $X_i$ in the graph $\graph$ denoted by $\vect{PA}^\graph_i \subset V(\graph)$. Specifically,
\begin{align}
\label{eq:dgp}
    X_i = f_i(\vect{X}_{\vect{PA}^\graph_i}, U_i),\; \text{ for some } f_i \in \Fcal \quad \forall i \in [d] ,
\end{align}
where $[d] \coloneqq \{1, \ldots, d\}$ and $\vect{X}_{\vect{C}}$ is the subvector of $\vect{X}$ formed by selecting elements indexed by $\vect{C}$.  
In this text, the entailed observational distribution $P_\vect{X}$ is often replaced with $P_\scm$ or even $P$ for brevity.
Refer to Appendix~\ref{appx:notation} for a formal discussion on SCMs and their assumptions.

\xhdr{Order Learning v.s. Structure Learning} 
Classic Bayesian network results confirm that causal structures can only be identified up to their Markov equivalence class, represented by a graph with undirected edges~\citep{spirtes2000causation}. 
Knowing the topological ordering of the causal graph will, in turn, allow us to infer the direction of the edges and accurately determine the underlying causal structure. This paper mainly focuses on learning the causal ordering. Learning the ordering is generally a simpler task since the space of orderings is strictly smaller than that of DAGs. Moreover, we can prove that order discovery can be achieved by strictly fewer assumptions than structure discovery. We refer the reader to \cref{appx:order_identifiability} for a concrete statement of such results and a detailed discussion on the benefits of order discovery. 

\xhdr{Fixed-Order Autoregressive Normalizing Flows} Normalizing flows are a class of generative models that produce the probability distribution of data using the change-of-variables formula~\citep{rezende2015variational}. With a slight abuse of notation, flows model the data distribution by parameterizing a diffeomorphic mapping $\vect{T}: \R^d \to \R^d$ from a base distribution $P_{\vect{U}}(\cdot)$ to the data distribution $P_\vect{X}(\cdot)$. Using change-of-variables, if $\vect{J}$ is the Jacobian of $\vect{T}^{-1}$, the density $P_\vect{X}$ can be directly computed as follows: 
\begin{align}\label{eq:change-var}
P_\vect{X}(\vect{x}) = P_\vect{U}(\vect{u}) \cdot \left| \det \vect{J}(\vect{x}) \right|, \text{ where } \vect{u} = \vect{T}^{-1}(\vect{x}).
\end{align}

Autoregressive normalizing flows (ANFs)~\citep{kingma2016improved} make the dependencies in the transform $\vect{T}$ ordered, and in turn, the Jacobian becomes upper-triangular, leading to tractable determinant computation for training. Although the final transform $\vect{T}$ is often obtained by stacking multiple autoregressive transforms with random orderings for enhanced expressiveness, one can fix the ordering to a specific permutation $\pi$ of the covariates to maintain the order as in SCMs~\citep{khemakhem2021causal}. Formally, if we assume $\pi$ to be a topological ordering of an SCM graph, then one can rewrite the link function $f_i$ in \eqref{eq:dgp} to comply with the flow notation: a deterministic mapping from \textit{all} the exogenous noises earlier in the ordering $\pi$ to $\vect{X}_i$. Specifically, if we write the transform $\vect{T}$ as $d$ one-dimensional mappings, the flow-induced distribution complies with the following format, resembling the data-generating SCM:
\begin{align}
X_{\pi(i)} = {T}_{\pi(i)}(U_{\pi(1)}, U_{\pi(2)}, \hdots, U_{\pi(i)}), \quad \forall i \in [d].
\end{align}

Contrary to \cite{khemakhem2021causal}, who focus on \emph{affine} ANFs primarily in bivariate cases, this paper presents a general and scalable framework for multivariate settings.

\xhdr{Permutation Learning} Learning a valid causal order can be framed as a search over the discrete space of all $d\times d$ permutation matrices denoted by $\BirkhoffVertices_d$, the vertices of the Birkhoff polytope visualized in \autoref{fig:main_figure}. Finding a valid permutation in this super-exponential space becomes intractable as $d$ grows. {One conventional parameterization of the permutation matrices is obtained using the matrix $\StructureBelief \in \R^{d \times d}$ and the matching function $M: \R^{d \times d} \to \BirkhoffVertices_d$ defined below}:
\begin{align}
M(\StructureBelief) := \argmax_{\HardPermutation \in \BirkhoffVertices_d} \langle \HardPermutation, \StructureBelief \rangle_F, \label{eq:matching-func}
\end{align}
where $\langle \cdot, \cdot \rangle_F$ is the Frobenious inner product. {Instead of solving \cref{eq:matching-func} for a deterministic value of $\StructureBelief$, we can parameterize a Boltzmann distribution over permutations, where the energy of a permutation $\HardPermutation \in \BirkhoffVertices_n$ is equal to $\langle \StructureBelief, \HardPermutation \rangle_F$. As illustrated heuristically in \citet{hazan2013sampling, tomczak2016some}, the Gumbel Matching distribution $\GumbelMatchingDistribution(\StructureBelief)$ can be used to approximate unbiased samples from the true Boltzmann distribution: Any Sample from this distribution is generated by first sampling $\epsilon \in \R^{d \times d}$ — a matrix of standard i.i.d. Gumbel noise — and passing $\StructureBelief + \epsilon$ through the matching function $M$. Since the matching function itself is not differentiable, prior studies introduce the Sinkhorn function $S: \R^{d \times d} \to \R^{d\times d}$ that iteratively normalizes rows and columns of its input matrix, resulting in a continuous relaxation of the matching function. In particular, one can replace $M(\StructureBelief)$ with $S(\StructureBelief/\tau)$ for a very small scaler $\tau$. The Sinkhorn function is proven to converge to the matching function as $\tau \to 0$, unlocking a gradient-based optimization called the Gumbel-Sinkhorn method ~\citep{mena2018learning}. In this work, rather than resorting to the Sinkhorn function as a differentiable approximation of the matching function, we use the exact probability values of the permutations sampled from $\GumbelMatchingDistribution$ as we discuss in \cref{sec:permutation-learning}.}

%% file: sections/related.tex
\section{Related Work}
Structure learning is a rich and expansive area of research that can significantly aid researchers in identifying the underlying causal structures. A comprehensive review of such techniques can be found in \cite{vowels2022d}. Our emphasis is on order-based structure learning \citep{verma1990causal,bouckaert1992optimizing,singh1993algorithm,friedman2003being,scanagatta2015learning,park2017bayesian,ruiz2022sequentially}, which inherently produce acyclic structures, consistent with causal DAGs. At its core, causal structure learning is a combinatorial search problem. Previous work has considered greedy hill-climbing search~\citep{teyssier2012ordering}, restricted maximum likelihood estimation (CAM)~\citep{buhlmann2014cam}, as well as more recent approaches such as sparsest permutation learning~\citep{raskutti2018learning,solus2021consistency,lam2022greedy}, and reinforcement learning~\citep{wang2021ordering} to tackle this problem. Recently, the score-matching algorithm (SCORE) has been proposed to find the correct orderings of variables in ANMs by computing the Hessian of the marginal log-likelihoods for each variable and choosing the one with the lowest variance as a sink node~\citep{rolland2022score,sanchez2022diffusion}. However, for more general model classes such as LSNMs, the Hessian of the leaf nodes does not necessarily remain constant, rendering the described method infeasible.

Another direction involves formulating the search over topological orderings as an end-to-end differentiable optimization. BCDNets~\citep{cundy2021bcd}, and VI-DP-DAG~\citep{charpentier2022differentiable} model the orderings as latent variables and use variational inference to approximate the posterior over orderings. These methods map each topological ordering to a vertex in the convex hull of permutation matrices, known as the Birkhoff polytope, and utilize the Gumbel-Sinkhorn approximations and permutation learning tricks to relax the discrete permutation space~\citep{bengio2013estimating, mena2018learning}.
More recently,~\citet{zantedeschi2023dag} proposed ``DAGuerreotype'' that considers parameterizing the permutation vectors by assigning a score to each node, with higher scores indicating higher ranks in the ordering. The SparseMap operator~\citep{niculae2018sparsemap} relaxes their structured search problem and makes it differentiable.
While these methods have been showcased in real-world settings without explicit assumptions, they lack theoretical identifiability justifications for the inferred causal orderings. In fact, the mean squared error loss functions employed by these methods can be viewed as a special instance of our flow-based log-likelihood loss, given a Gaussian prior with fixed variance, and their methods should only provably work on ANM-compliant data-generating processes.

Given causal identification is impossible under no assumptions, several works have proposed loose assumptions to identify cause-effect directions. These include restricting causal mechanisms to non-linear models with additive noise~\citep{peters2014causal,chicharro2019conditionally} or non-Gaussian linear models~\citep{shimizu2006linear}, assuming non-parametric constraints on the variance or conditional entropy of exogenous noises~\citep{peters2014identifiability,ghoshal2018learning, chen2019causal,gao2020polynomial,gao2021efficient} and additional information like non-stationary time-series data~\citep{monti2020causal}. Recent work has demonstrated causal direction identifiability for causal models with heteroscedastic noises (also known as location-scale noise models)~\citep{khemakhem2021causal,strobl2022identifying,immer2022identifiability}. {However, their proposed structure learning methods are mainly applicable to bivariate settings. This work builds on their theoretical results to provide a provable and scalable causal structure learning for the multivariate case.} 

%% file: sections/method.tex
\section{\method\ Framework} 
\label{sec:method}
In simple scenarios involving two variables, a likelihood ratio test is often employed to determine the cause-effect direction, where the direction with the highest likelihood value identifies the correct ordering~\citep{khemakhem2021causal, immer2022identifiability}. A natural extension to multivariate settings is to pick pairs of covariates and run the test to determine which comes earlier in the causal ordering. But, the characteristics of the link function that govern the entire SCM break when we limit ourselves to a subset of covariates, and therefore, such pairs may become non-identifiable (See \cref{appx:bivar-multivar}). Our experimental results in \cref{sec:experiments} confirm this in practice.

We can extend the bivariate likelihood ratio test in \citet{khemakhem2021causal} to multivariate settings by defining autoregressive flows as our likelihood model, performing a maximum likelihood estimation-based search over all possible orderings, and finding the order that has the highest estimated likelihood. Formally, suppose we are given an observational dataset $\data = \{\vect{x}^{(i)}\}_{i=1}^N$ that is sampled from an SCM $\scm$. For any ordering $\vect{P} \in \BirkhoffVertices_d$, an ANF parameterized by $\theta \in \Theta$ takes a data point $\vect{x}^{(i)}$ as input and computes the likelihood $P_{\vect{X}}(\vect{x}^{(i)}; \vect{P}, \theta)$ through change-of-variables. With that in mind, the ordering that obtains the maximum likelihood, $\vect{P}^\star$, can be defined as follows:
\begin{align} \label{eq:max-log-like}
     & \vect{P}^* \coloneqq \argmin_{\vect{P} \in \BirkhoffVertices_d} \ell^*_{N}(\vect{P}),\ \text{ where }\
     \ell^*_{N}(\vect{P})  \coloneqq \min_{\theta \in \Theta}  \frac{1}{N} \sum_{i=1}^N - \log P_{\vect{X}}(\vect{x}^{(i)}; \vect{P}, \theta). \nonumber
\end{align}
However, even in infinite data, $P^\star$ does not necessarily correspond to the correct causal ordering. This is because, when attempting to parameterize the data-generating process using a generative model such as an ANF, overly expressive models can result in non-identifiability, i.e., there can be multiple ANFs with contradictory topological orderings while entailing the same distribution as the observational data. 
Therefore, the model complexity of the generative ANF should match the assumptions required for identifiability. Recently, \citet{immer2022identifiability,strobl2022identifying} have proved identifiability for a subset of location-scale noise models (LSNMs) by defining a set of extra conditions on the model class, which we call \textit{restricted} LSNMs. Those theoretical results, however, do not distinguish between the \textit{data complexity class} and \textit{the search model class}. By data complexity class, we mean the complexity of the true data-generating process, while the search model class refers to the class of parametric likelihood models we use to find the correct ordering. In particular, \citet{khemakhem2021causal} showed that the model class of \textit{affine} ANFs is equivalent to the data complexity class of LSNMs. Therefore, we may want to use affine ANFs as our parametric model to run the aforementioned likelihood-based approach. However, affine ANFs do not generally satisfy the extra conditions required for restricted LSNMs. Due to this mismatch between the model and data classes, we cannot apply the mentioned theoretical results to provide the correctness of a likelihood-based search. Therefore, our following proposition provides a stronger identifiability result than the previous work by \underline{not} limiting the search model class to restricted LSNMs.
\begin{restatable}{proposition}{anfcorrectprop} \label{appx-prop:ordering-loss}
    Consider an SCM $\scm$ from the family of restricted LSNMs and denote $\Theta$ as the set of all "conventional" affine ANFs w.r.t. $\scm$. For each permutation $\pi$, let $\ell^*_\infty(\pi)$ be the minimum expected negative log-likelihood under this model. That is,
    \begin{align}
        \ell^*_{\infty}(\pi) \coloneqq \min_{\theta \in \Theta} \E_{\vect{x} \sim P_{\scm}}\left[- \log P_{\vect{X}}(\vect{x}; \pi, \theta)\right].
    \end{align}
    Denote $\Pi_\graph$ as the set of all valid causal orderings of $\graph$. Then, $\forall \pi \in \Pi_{\graph_\scm}, \bar{\pi} \not\in \Pi_{\graph_\scm}$, it holds that $\ell^*_\infty(\pi) < \ell^*_\infty(\bar{\pi})$. In other words, the expected negative log-likelihood of all valid orderings is strictly smaller than those of other orderings.
\end{restatable}
The "conventional" term in \cref{appx-prop:ordering-loss} refers to weak conditions on the ANF model, such as the continuity of the transform functions. We provide a detailed definition of restricted LSNMs and conventional ANFs and the proof in \cref{appx:proof-ordering-loss}. \cref{appx-prop:ordering-loss} theoretically backs up the naive likelihood-based search described above by proving that the correct orderings are identifiable for the LSNM data complexity class, even if we use the search model class of ANFs that are more expressive than restricted LSNMs.

Still, there remain concerns around scalability. 
Defining a separate ANF model for each order $\vect{P}$ to compute $P_{\vect{X}}(\cdot; \vect{P}, \theta)$ requires a super-exponentially large number of models and/or parameters.
Furthermore, finding $\vect{P}^*$ hinges on a search over the space of all orderings, which is a discrete set that grows super-exponentially
as $d$ increases. 
This renders the search intractable and non-differentiable, precluding the possibility of using gradient-based optimization. 
Even for moderately large structures, these two problems pose a serious challenge in structure learning for real-world scenarios. In the following sections, (i) we present our masked ANF architecture that can simultaneously model data-generating processes with different orderings, and (ii) present a novel permutation learning approach which is both scalable and differentiable and can thus be combined with the ANF learning framework.

\xhdr{Remark} As described in \cref{sec:problem-set}, we mainly focus on learning the causal ordering in \method. However, we note that once the correct ordering $\vect{P}^*$ is estimated, we can uncover the causal DAG structure by deploying any algorithm to find the undirected causal skeleton \footnote{Specifically, we consider PC-KCI and sparse-regression techniques for this phase. More details are provided in \cref{sec:experiments}.} and overlaying the optimal order onto this structure to find the final DAG. 
\subsection{Masked Flow Ensembles for Modelling Different Orderings}
\label{subsec:lsnm-flow}
For any input value $\vect{u}$, a transformation $\vect{T}(\vect{u})$ in an ANF is computed using a set of values that are themselves obtained by passing $\vect{u}$ through one or a combination of autoregressive neural networks with learnable parameters.
A transform
$\vect{T}(\cdot; \xi): \R^d \to \R^d$ is parameterized by a set of \textit{non-learnable} values $\xi \in \Xi$ which are in turn outputs of a \textit{learnable} function
$\vect{\Psi}(\cdot; \theta): \R^d \to \Xi$ parameterized by $\theta$.

In affine ANFs $\vect{\Psi}(\cdot, \theta)$ maps $\vect{u}$ to two $d$-dimensional outputs vectors using two autoregressive models $\vect{\scalefunc}(\cdot; \theta)$ and $\vect{\locfunc}(\cdot; \theta)$ -- i.e., 
$\vect{\Psi}(\vect{u}; \theta) \coloneqq (\vect{\scalefunc}(\vect{u}; \theta), \vect{\locfunc}(\vect{u}; \theta))$. Subsequently, $\vect{T}(\vect{u}; \vect{\Psi}(\vect{u}; \theta)) \coloneqq \vect{\scalefunc}(\vect{u}; \theta) \odot \vect{u} + \vect{\locfunc}(\vect{u}; \theta)$ indicates an affine transform where $\odot$ is an elementwise multiplication. 

Like \citet{germain2015made}, we mask the weights of neural networks modelling $\vect{\Psi}(\cdot; \theta)$ to ensure their inputs and intermediate representations form an autoregressive structure and that each output entry only depends on its preceding input entries. 
In turn, $\vect{\Psi}(\vect{u}; \theta)$ can be defined using a collection of masked feed-forward autoregressive network that is faithful to the order $\vect{P} \in \BirkhoffVertices_d$:
\begin{align} \label{eq:anf-arch} 
\sigma\left( \hat{\vect{M}}_\vect{P} \odot W^{(L)} \sigma \left( \vect{M}_\vect{P} \odot W^{(L-1)}  \sigma \left(\ldots  \vect{M}_\vect{P} \odot W^{(1)} \vect{u} \right)\right)\right),
\end{align}
where $\{W^{(l)}\}_{l=1}^L$ are weight matrices that collectively form $\theta$, $\sigma: \R^d \to \R^d$ is a non-linear activation, and $\vect{M}_{\vect{P}}, \hat{\vect{M}}_{\vect{P}}$ are block lower triangular masking matrices that align the model with the autoregressive order $\vect{P}$. The former allows self-dependencies with non-zero diagonals, and the latter prevents them with zero ones, ensuring the autoregressiveness of the network (see \cref{appx:arch} for more details).

We can change the underlying order of dependencies in our ANF by plugging in different masks associated with the desired ordering. This enables sharing parameters across different orderings by holding $\theta$ constant and plugging different masking matrices.
If $\vect{M}_\idenperm$ denotes a lower triangular all-ones matrix, we can define appropriate masks by replacing $\vect{M}_{\vect{P}}$ and $\hat{\vect{M}}_{\vect{P}}$ with $\vect{P}^\top \vect{M}_\idenperm \vect{P}$ and $\vect{P}^\top (\vect{M}_\idenperm - \vect{I}) \vect{P}$ respectively.

Training $\theta$ to maximize likelihood over the dataset $\data$ by feeding different permutation matrices $\vect{P}$ enables the model to learn data-generating processes for all the input orderings simultaneously. As we will see in the next part, when combining this method with our permutation learning framework, the model reliably captures the negative log-likelihood $\ell_N^*(\vect{P})$ for permutation $\vect{P}$ that is more likely to be sampled from $\GumbelMatchingDistribution(\StructureBelief)$ at any given time.

\subsection{Permutation Learning} \label{sec:permutation-learning}
{The previous section described our procedure as finding a permutation $\vect{P}^*$ that minimizes $\ell^*_N(\vect{P})$.
Using the Boltzmann distribution introduced in \cref{sec:problem-set} sidesteps the issue of non-differentiability by parameterizing a distribution over the set of all orderings using the matrix $\StructureBelief$.
In turn, the expected value of $\ell^*_N(\vect{P})$ for $\vect{P}$ being sampled from this distribution can serve as a differentiable loss to minimize and set $\StructureBelief$ such that the distribution collapses onto $\vect{P}^*$. Formally, we denote this loss as}
\begin{align} \label{eq:proxy-loss}
\ProxyLoss(\StructureBelief) = \sum_{\HardPermutation_\pi \in \BirkhoffVertices_d} \alpha(\HardPermutation_\pi) \cdot \ell^*_N(\pi),
\end{align}
{where $\alpha(\vect{P}) \coloneqq \frac{\exp{\langle \StructureBelief \cdot \HardPermutation \rangle_F}}{Z(\StructureBelief)}$ is the probability mass associated with $\vect{P}$ and $Z(\StructureBelief)$ is the normalizing factor.
\cref{fig:main_figure} illustrates this loss function. In particular, the permutations close to 
$\StructureBelief$ are more likely to be sampled, and $\StructureBelief$ can converge to a value such that the Boltzmann distribution only generates the permutation $\vect{P}^*$, minimizing $\ProxyLoss(\StructureBelief)$. }

{Evaluating the normalization factor in \cref{eq:proxy-loss} requires summing over all the $d!$ permutations; therefore, although the proxy loss $\ProxyLoss$ solves the non-differentiability issue, its computation is intractable. A common strategy (see \cref{sec:problem-set}) to circumvent this intractability is to approximate the Boltzmann distribution using the Gumbel-Matching distribution $\GumbelMatchingDistribution(\StructureBelief)$, which entails the following loss function that can be approximated via Monte-Carlo estimation:}
\begin{equation} \hat{\ProxyLoss}^{\GumbelMatchingDistribution}(\StructureBelief) = \E_{\HardPermutation_\pi \sim \GumbelMatchingDistribution(\StructureBelief)} \left[ \ell^*_N(\pi)\right].
\end{equation}
Sampling from the Gumbel-Matching distribution involves sampling Gumbel noise $\epsilon$ and applying the matching function on $\StructureBelief + \epsilon$ which is non-differentiable. 
The similarity between the Sinkhorn and Matching functions in combination with straight-through gradient estimators~\citep{bengio2013estimating} can used to perform gradient-based learning. 
In the forward pass, $M(\StructureBelief + \epsilon)$ is used, and in the backward pass, $M(\StructureBelief + \epsilon)$ is replaced by the differentiable Sinkhorn function $S((\StructureBelief + \epsilon)/\tau)$ with a small $\tau$. 

{Although replacing the matching operator in tandem with the gradient straight-through estimator is commonplace in permutation learning~\citep{mena2018learning, kool2019stochastic, charpentier2022differentiable}, we introduce a novel technique to optimize \cref{eq:proxy-loss}. We draw a set of unbiased samples from $\GumbelMatchingDistribution(\StructureBelief)$ and remove repeated entries to obtain a tractable set of ``most likely'' permutations $\mathcal{H} = \{\HardPermutation^{(k)}\}_{k=1}^{|\mathcal{H}|}$. We then approximate the probability masses $\alpha(\vect{P})$ as zero if $\vect{P} \notin \mathcal{H}$ and as follows otherwise:}
\begin{equation}
    \hat{\alpha}(\vect{P}) \coloneqq \frac{\exp\langle \vect{P}, \vect{\StructureBelief}\rangle_F}{\sum_{\vect{P}' \in \mathcal{H}} \exp\langle \vect{P}, \vect{\StructureBelief}\rangle_F}
\end{equation}
{Using the values of $\hat{\alpha}$ above, we can define an approximation of $\ProxyLoss(\StructureBelief)$ in \cref{eq:proxy-loss} by $\widehat{\ProxyLoss}(\StructureBelief) := \sum_{\HardPermutation \in \mathcal{H}} \hat{\alpha}(\HardPermutation) \cdot \ell^*_N(\HardPermutation)$, which is both differentiable and tractable.}
{To combine this loss into our architecture, we replace $\ell^*_N(\vect{P})$ with the average negative log-likelihood the ANF obtains on a batch of observational data while the ANF dependencies are masked using matrix $\vect{P}$. We then concurrently optimize the ANF parameters $\theta$ and permutation parameters $\StructureBelief$ using alternating optimization on the proxy loss $\widehat{\ProxyLoss}$. At the end of the training, the most frequently generated permutation is considered as the correct ordering. See \cref{appx:algo} for more details.}

%% file: sections/experiments.tex
\section{Experiments}
\label{sec:experiments}
\begin{table*}
  \caption{\small Comparison of \method\ with baselines on Sachs and SynTReN datasets using CBC, SHD, and SID metrics. Equipped with the Gumbel-Top-$k$ trick for order learning, ours obtains state-of-the-art SHD and competitive results on other metrics. (For the VI-DP-DAG baseline, $^1$ uses the Gumbel-Sinkhorn, and $^2$ uses the Gumbel-Top-$k$ approach {\citep{kool2019stochastic}}).}
  \label{tab:metrics-sachs-syntren}
  \begin{adjustbox}{width=\linewidth,center}
  \tiny
  \begin{tabular}{lcccccc}
    \toprule
    \multirow{2}{*}{} & \multicolumn{3}{c}{\textbf{Sachs} ~(mean $\pm$ std)}& \multicolumn{3}{c}{\textbf{SynTReN} ~(mean $\pm$ std)} \\
    \cmidrule(lr){2-4} \cmidrule(lr){5-7}
    & \textbf{CBC} & \textbf{SHD} & \textbf{SID} & \textbf{CBC} & \textbf{SHD} & \textbf{SID} \\
    \midrule
    \method\ {\scalebox{0.9}{(PC-KCI test)}} & 0.18 $\pm$ 0.11 & \textbf{10.4 $\pm$ 0.89}  & {46.4 $\pm$ 6.77} & \textbf{0.21 $\pm$ 0.17} & \textbf{32.0 $\pm$ 2.9} & 161.0 $\pm$ 58.9 \\
    \method\ {\scalebox{0.9}{(sparse-regression)}} & 0.18 $\pm$ 0.11 & \textbf{10.4 $\pm$ 0.89} & {46.4 $\pm$ 6.77} & \textbf{0.21 $\pm$ 0.17} & 34.3 $\pm$ 3.4 & \textbf{122.9 $\pm$ 57.1} \\
    \midrule
    CAM & 0.41 $\pm$ 0.0 & 12 $\pm$ 0.0 & 55 $\pm$ 0.0 & 0.33 $\pm$ 0.14 & 41.7 $\pm$ 7.1 & 139.6 $\pm$ 36.1 \\
    SCORE & 0.47 $\pm$ 0.0 & 12 $\pm$ 0.0 & 45 $\pm$ 0.0 & 0.38 $\pm$ 0.10 & 37.5 $\pm$ 4.4 & 197.1 $\pm$ 67.1 \\
    VarSort & 0.47 $\pm$ 0.0 & 12 $\pm$ 0.0 & 45 $\pm$ 0.0 & 0.51 $\pm$ 0.21 & 46 $\pm$ 9.9 & 187.4 $\pm$ 81.1 \\
    DAGuerreo & \textbf{0.13 $\pm$ 0.03} & 20.4 $\pm$ 0.9 & 48.8 $\pm$ 2.1 & 0.40 $\pm$ 0.16  & 74.1 $\pm$ 11.5 & 159.4 $\pm$ 69.4 \\
    VI-DP-DAG$^1$ & 0.34 $\pm$ 0.11 & 29 $\pm$ 1.8 & 46.2 $\pm$ 3.3 & 0.6 $\pm$ 0.21 & 142.1 $\pm$ 8.5 & 147.1 $\pm$ 36.8 \\
    VI-DP-DAG$^2$ & 0.48 $\pm$ 0.12 & 33 $\pm$ 2.9 & \textbf{41.6 $\pm$ 6.2} & 0.58 $\pm$ 0.17 & 137.2 $\pm$ 9.4 & 152.6 $\pm$ 40.4 \\
    bi-LSNM  & 0.59 $\pm$ 0.0 & 19 $\pm$ 0.0 & 59 $\pm$ 0.0 & 0.46 $\pm$ 0.17 & 49.5 $\pm$ 7.6 & 170.1 $\pm$ 75.6 \\
    GraN-DAG & - & 13 $\pm$ 0.0 & 47 $\pm$ 0.0 & - & 34.0 $\pm$ 8.5 & 161.7 $\pm$ 53.4 \\
    \bottomrule
  \end{tabular}
  \end{adjustbox}
\end{table*}

\xhdr{Organization} (i) We first assess \method\ in real-world and semi-synthetic scenarios to highlight the significance of minimizing prior assumptions, such as additive noises. We utilize the Sachs~\citep{sachs2005causal} and SynTReN~\citep{van2006syntren} datasets for these experiments, demonstrating competitive results in terms of structural hamming distance (SHD) and structural intervention distance (SID) \citep{peters2015structural}.
(ii) We compare our permutation learning approach to the previous baselines in the literature on a diverse set of appropriately specified synthetic data, particularly those adhering to identifiable distributions such as ANMs and LSNMs.
(iii) We provide empirical evidence that causal ordering, without a full structure, can effectively estimate interventional distributions, highlighting our model's ability for causal effect estimation with observational data. See Appendix \ref{appx:experimental-details} for a detailed description of our experiments and the hyperparameters.

\xhdr{Dataset}
We consider the real-world genetic dataset Sachs, which contains $N = 853$ observations of $d = 11$ proteins from human immune system cells. Moreover, to show \method's scalability to moderately large graph structures, we evaluate $10$ datasets from SynTReN, each featuring $N = 500$ semi-simulated gene expressions across $d = 20$ nodes, with $20$ to $25$ edges. We also include a comprehensive synthetic benchmark of various-sized DAGs with different graph structures. To design the link functions for these synthetic datasets, we either sample ``nonparametric'' functions from a Gaussian process to sidestep any parametric assumption \citep{zhu2019causal}, or we directly specify them by randomly generating the parameters of a parameterized function family, indicated by ``parametric''. 
Our synthetic benchmark is thorough and offers a robust testing environment for causal discovery (see Appendix \ref{appx:synthetic-benchmark}).
\begin{table*}[t]
\centering
\caption{\small Comparison of $\cbc$ metric (mean $\pm$ std) for different graph structures (chain, full-graph, Erdős–Rényi) and different seeds. {In addition to our original implementation of \method, we also implement the Gumbel-Sinkhorn method~\citep{mena2018learning} as another permutation learning module as an ablation against \method} We run all the baselines on datasets with $d \in \{3, 4, 5, 6\}$ to show how relaxing the ANM assumption helps structure learning on different synthetic datasets. }
\label{tab:synthetic}
\resizebox{\linewidth}{!}{%
\footnotesize
\begin{tabular}{lccccccc}
\toprule
& \textbf{Linear} & \multicolumn{2}{c}{\textbf{Linear Laplace}} & \multicolumn{2}{c}{\textbf{Nonparametric}} & \multicolumn{2}{c}{\textbf{Nonlinear Parametric}} \\
\cmidrule(lr){3-4} \cmidrule(lr){5-6} \cmidrule(lr){7-8} 
& \textbf{Gaussian} & \textbf{Affine} & \textbf{Additive} & \textbf{Affine} & \textbf{Additive} & \textbf{Affine} & \textbf{Additive} \\
\midrule
\method  & $0.46 \pm 0.27$ & $\mathbf{0.22 \pm 0.20}$ & $\mathbf{0.16 \pm 0.18}$ & $\mathbf{0.12 \pm 0.14}$ & $0.13 \pm 0.15$ & $\mathbf{0.15 \pm 0.18}$ & $\mathbf{0.20 \pm 0.19}$ \\
\method\ {\scalebox{0.9}{(Gumbel-Sinkhorn)}} & $0.48 \pm 0.28$ &  $0.42 \pm 0.28$ & $0.46 \pm 0.30$ &  $0.29 \pm 0.22$ & $0.37 \pm 0.29$ &$0.38 \pm 0.24$  & $0.42 \pm 0.28$ \\
\midrule
CAM & $0.44 \pm 0.34$ &   $0.86 \pm 0.19$ &  $0.67 \pm 0.38$ & $0.27 \pm 0.25$ &  $0.02 \pm 0.07$ & $0.83 \pm 0.27$ &  $\mathbf{0.19 \pm 0.25}$ \\
VI-DP-DAG  & $0.59 \pm 0.34$ & $0.67 \pm 0.3$ & $0.4 \pm 0.28$  & $0.42 \pm 0.25$& $0.42 \pm 0.26$ & $0.65 \pm 0.23$  & $0.51 \pm 0.25$ \\
DAGuerreo &  $0.4 \pm 0.28$ & $0.81 \pm 0.21$ & $0.5 \pm 0.32$   &  $0.27 \pm 0.22$ &$0.14 \pm 0.19$  &$0.74 \pm 0.31$  & $0.44 \pm 0.30$  \\
SCORE  & $0.53 \pm 0.29$  & $0.68 \pm 0.31$ & $0.73 \pm 0.3$   & $\mathbf{0.1 \pm 0.17}$ & $\mathbf{0.00 \pm 0.00}$ &  $0.78 \pm 0.35$  & $0.44 \pm 0.40$ \\
VarSort & $0.39 \pm 0.16$  & $0.52 \pm 0.24$ & $0.47 \pm 0.23$ & $0.36 \pm 0.25$ & $0.17 \pm 0.19$  & $0.66 \pm 0.43$ & $0.49 \pm 0.37$ \\
biLSNM  & $0.54 \pm 0.22$  & $0.59 \pm 0.19$ & $0.56 \pm 0.17$ &  $0.56 \pm 0.21$& $0.56 \pm 0.21$  & $0.57 \pm 0.22$  & $0.57 \pm 0.22$ \\
\bottomrule
\end{tabular}%
}
\end{table*}

\xhdr{Baselines}
We consider order-based methods, including CAM \citep{buhlmann2014cam}, DAGuerreo \citep{zantedeschi2023dag}, SCORE \citep{rolland2022score}, and VI-DP-DAG \citep{charpentier2022differentiable}, as well as GraN-DAG \citep{lachapelle2019gradient}, a continuous optimization method that achieves state-of-the-art results on the Sachs and SynTReN datasets.
Additionally, we use the likelihood ratio testing method proposed by \citet{immer2022identifiability} for bivariate location-scale noise models (bi-LSNM), adapting it to our multivariate setting, wherein for each pair of nodes, we apply the test and sort the nodes based on the pairwise comparisons. We do not use the method proposed by \citet{khemakhem2021causal} as our baseline due to its expensive running time --- It would require training multiple autoregressive flows for all the pairs of random variables in the data.
\cite{reisach2021beware} posit that the marginal variances of simulated data alone can yield the correct causal ordering. We thus standardize the marginals in our simulations and compare them against ``VarSort'' which directly infers the causal ordering by sorting the marginal variances. See Appendix \ref{appx:baselines} for more details on our baselines.

\xhdr{Order-Based Evaluation}
Beyond well-established metrics SHD and SID that directly compare causal graphs, we introduce the causal backward count (CBC) as a specialized metric for evaluating causal orderings. This metric leverages the concordance index over causal pairs. Since order-based methods comprise two phases --- determining the causal skeleton and estimating the ordering --- CBC specifically gauges the quality of the ordering. This distinction clarifies if a model's shortcomings originate from the order learning phase.
For an estimated ordering $\hat{\pi}$, CBC is defined as:
\begin{equation} \label{eq:cbc}
\cbc_\graph(\hat{\pi}) = \scalebox{1.5}{$\Sigma$}_{(i,j) \in E(\graph)} {\small \II(\hat{\pi}(i) > \hat{\pi}(j)) / |E(\graph)|} ,
\end{equation}
where $E(\graph)$ represents the edges in the ground-truth causal graph and $\II(\cdot)$ is the indicator function.

\xhdr{Real-World and Semi-Synthetic Results}  
\autoref{tab:metrics-sachs-syntren} presents a comparative analysis of \method\ with other baselines on Sachs and SynTReN, employing different evaluation metrics. 
We use two standard methods to compute the causal skeleton: sparse regression after order learning \citep{buhlmann2014cam}, and PC with kernel-based conditional independence (PC-KCI) test \citep{zhang2012kernel}.
\method\ surpasses most baselines in terms of CBC, SHD, and SID, achieving state-of-the-art results on SHD.
To account for randomness, we run each method for $5$ times and report each metric's mean and standard deviation. Even though \method\ outperforms many baselines on the larger SynTReN dataset and demonstrates better scalability, it is interesting to see that there is still significant room for improvement as an empty graph would achieve an SHD in the range of $[20, 25]$.

\xhdr{Synthetic Results}
\autoref{tab:synthetic} compares \method~with other order-based baselines on simulated datasets using the CBC metric.
These results confirm the significance of implicit assumptions such as ANM in other baselines.
As a sanity check, note that all models perform comparably and poorly on the linear Gaussian case (first column), a provably non-identifiable case. While \method\ matches baseline performances on ANM benchmarks (except for CAM and SCORE outperforming in nonparametric settings), its performance markedly improves on affine datasets. It is worth mentioning that the nonparametric experiments do not necessarily comply with the conditions described in \citet{strobl2022identifying} for identifiability as they contain non-invertible link functions. This can explain the mixed results we observe in the nonparametric column of \cref{tab:synthetic}. In \cref{sec:method}, we emphasized the flexibility \method~offers in choosing the base distribution for likelihood computation. To validate this, we evaluate it on a synthetic dataset with exogenous noise following a Laplace distribution and linear link functions---an empirically identifiable dataset \citep{khemakhem2021causal}. While most methods struggle, \method\ incorporates the prior into the ANF architecture, leading to significantly enhanced performance, evident in the second and third columns.
Last but not least, we evaluate the permutation learning module of \method\ described in \cref{sec:permutation-learning} by comparing it to a modification of \method\ that uses the Gumbel-Sinkhorn method with the straight-through gradient estimator (second row of \autoref{tab:synthetic}). The results showcase the superiority of our permutation learning module and motivate its potential advantage as a standalone algorithm for searching over any discrete search space, which we leave as future work. For the larger graph sizes $d \in \{10, 25\}$, see Appendix \ref{appx:synthetic-benchmark}.

\begin{figure}
    \centering
\includegraphics[width=0.45\textwidth]{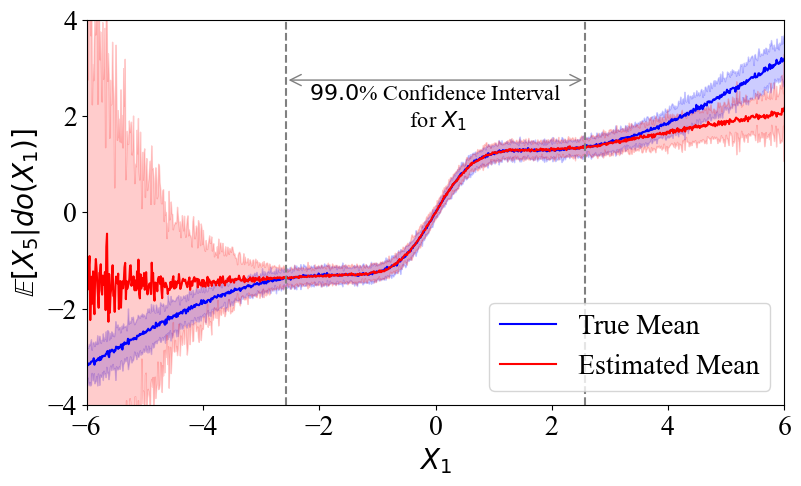}
    \caption{\small Estimating the interventional expected value $\E[X_5 | do(X_1)]$ using \method\ in a full causal graph from $X_1$ to $X_5$. The estimated value matches the true expectation in the $99\%$ confidence interval of the observational data.}
    \label{fig:intervention}
\end{figure}

\xhdr{Interventions} 
Only the causal ordering of a data-generation process (without access to the full DAG) can be sufficient for consistent estimation of interventional distributions~\citep{buhlmann2014cam}. Based on this, \method\ allows for sampling from interventional distribution by first learning the causal ordering from the input data and then harnessing the generative capabilities of ANFs for sampling. To empirically validate this, we extend the empirical setup for bivariate scenarios in \cite{khemakhem2021causal} into the multivariate domain.
In \autoref{fig:intervention}, we use \method\ to train an ANF on the data generated from a causal graph consisting of a full causal graph with five nodes, where the learned ordering is the same as the true one. We then estimate $\E[X_5 | do(X_1)]$ by performing interventions on the covariate $X_1$ across the x-axis. The figure demonstrates that the estimated mean value matches the true interventional expected value for $X_1 \in [-2.5, 2.5]$, which consists of the $99\%$ of the observational data. However, the estimation error intensifies as $X_1$ extends beyond this interval, likely due to the lack of such $X_1$ values in the observational data. See Appendix \ref{appx:interventions} for more experiments.

%% file: sections/conclusion.tex
\section{Conclusion \& Limitations}
We introduced \method, a likelihood-based scalable method for causal structure learning from observational data that relaxes prior assumptions on the data-generating process. \method\ : (i)  uses masked autoregressive normalizing flows as an ensemble model to keep track of the likelihood values for multiple topological orderings at once, and (ii) guides the search to find the correct causal ordering using a novel permutation learning trick. Our results on Sachs and SynTReN underscore the significance of addressing implicit assumptions in causal discovery. This paves the way for more accurate methodologies with practical applicability.

While our ANF framework is articulated in a broad context, our results are provable for the affine ANF setting where data adheres to LSNMs. Exploring more general scenarios can inform the development of structure learning techniques that embrace the ANF structure without particular model specifications. A potential extension involves integrating a non-linear layer post-transformation, accommodating post-non-linear models \citep{zhang2012identifiability}. 

Finally, \method~introduces added complexity compared to prior baselines with the ANM assumption. This can occasionally hinder performance. For instance, in the additive nonparametric synthetic experiments, where the ANM assumption holds, SCORE outperforms us, and on the Sachs dataset, DAGuerreo and VI-DP-DAG surpass our results in terms of CBC and SID. Hence, before deploying our framework in real-world applications, it is vital to pin down any potential drawbacks of removing the ANM constraints 
and identify scenarios where such removal is appropriate.

%% file: appendix.tex
\input appendices/prelim
\input appendices/proofs
\input appendices/algo
\input appendices/experiment_setup
\input appendices/interventions

%% file: appendices/prelim.tex
\section{SCM Assumptions and Identifying the Order}
\label{appx:assumptions}
\subsection{Assumptions in Causal Structure Discovery and Causal Minimality}
\label{appx:notation}

All the results discussed in the paper rely on the \textit{sufficiency} assumption, which rules out unobserved confounding, as well as \textit{causal minimality} that ensures the causal discovery problem is well-defined and has a \textit{unique} answer. Intuitively, if $X_i$ is a parent of $X_j$, it should have a discernible impact on $X_j$'s distribution. Otherwise, arbitrary edges of zero influence can make the answer to a causal discovery problem non-unique. Here, we formally define the causal minimality assumption for SCMs.
\begin{definition}[Causal Minimality]
    A SCM $\scm = (\graph, \Fcal, P_{\vect{U}})$ satisfies causal minimality if there is no other SCM $\bar{\scm} = (\bar{\graph}, \bar{\Fcal}, \bar{P}_{\vect{U}})$ with $\bar{\graph} \subset \graph$ such that $P_{\scm} = P_{\bar{\scm}}$.
\end{definition}
A more explicit assumption on the link functions that closely resembles causal minimality is the notion of non-constant SCMs, as detailed below. Before formally defining non-constant SCMs, we note that throughout the text for any subset $\vect{C} \subseteq V(\graph_\scm)$, we call $\vect{x}_\vect{C}$ a \textit{realization} of $\vect{X}_\vect{C}$ iff the density entailed from the probability $P_\vect{X}$ on $\vect{X}_\vect{C}$ is positive at $\vect{x}_\vect{C}$.
\begin{definition}[Non-constant SCMs] \label{def:non-constant}
    An $\scm = (\graph, \Fcal, P_{\vect{U}})$ is non-constant iff for each $i \in V(\graph)$ and $j \in \vect{PA}^\graph_i$, there exists $u_i\sim P_{U_i}(\cdot)$ and two realizations  $\vect{x}_{\vect{PA}^\graph_i} = (\vect{x}_{\vect{PA}^\graph_i \backslash{\{j\}}}, x_j)$ and $\vect{x}'_{\vect{PA}^\graph_i} = (\vect{x}_{\vect{PA}^\graph_i \backslash{\{j\}}}, x'_j)$ such that $x_j \neq x'_j$, and
    $$f_i(\vect{X}_{\vect{PA}^\graph_i} = \vect{x}_{\vect{PA}^\graph_i}, U_i=u_i) \neq f_{i}(\vect{X}_{\vect{PA}^\graph_i} = \vect{x}'_{\vect{PA}^\graph_i}, U_i=u_i).$$
\end{definition}
Though it might be counter-intuitive at first glance, non-constant and minimal SCMs are not the same: \emph{An SCM can be non-constant yet not minimal}. Examples where these two concepts differ are fairly pathological. In many situations, like continuous ANMs (refer to Proposition 17. of \citet{peters2014causal}) and LSNM (see Appendix \ref{appx:LSNM-minimality}), these two are equivalent. Specifically, the subsequent results indicate that a unique set of causal ordering can be identified even in cases where the causal DAG is non-unique. This means that we can even relax the causal minimality assumption for such model classes if our goal is only to find the true causal orderings (e.g., for understanding causal effects or modelling interventions).

\begin{lemma}\label{lemma:equivalence_non_constant}
    Assume a family of SCMs where causal minimality is equivalent to non-constant link functions. Then, there is a unique minimal SCM $\scm_0$ with graph $\graph_0$ s.t it is Markov w.r.t $P_\vect{X}(.)$ and for any other Markov SCM $\scm$ with graph $\graph_\scm$, we have that $\graph_0 \subseteq \graph_\scm$. 
\end{lemma}
\begin{proof}
We prove this by contradiction. We assume that for any SCM $\scm$ being Markov w.r.t. $P_\vect{X}(\cdot)$ such a unique graph does not exist and there are two \textit{different} Markov and causaly minimal SCMs $\scm_1$ and $\scm_2$ s.t. $\graph_{\scm_1} \subseteq \graph_{\scm}$ and $\graph_{\scm_2} \subseteq \graph_{\scm}$ and $\graph_{\scm_1} \neq \graph_{\scm_2}$. For any node $i$ with link function $f_i$ in $\scm$, define $\vect{A}_1 \coloneqq \vect{PA}_i^{\graph_{\scm}} \setminus \vect{PA}_i^{\graph_{\scm_1}}$ and $\vect{A}_2 \coloneqq \vect{PA}_i^{\graph_{\scm}} \setminus \vect{PA}_i^{\graph_{\scm_2}}$, and note that $X_i \indep \vect{X}_{\vect{A}_1} \mid \vect{X}_{\vect{PA}_i^{\graph_{\scm_1}}}$. The equivalence between non-constant SCMs and causal minimality implies that for any arbitrary realization $\vect{x}_{\vect{A}_1}$ we have that $f_i(\vect{x}_{\vect{A}_1}, \vect{X}_{\vect{PA}_i^{\graph_{\scm_1}}}, U_i) = f_i( \vect{X}_{\vect{PA}_i^{\graph_{\scm}}}, U_i)$. Similarly, $f_i(\vect{x}_{\vect{A}_2}, \vect{X}_{\vect{PA}_i^{\graph_{\scm_2}}}, U_i) = f_i( \vect{X}_{\vect{PA}_i^{\graph_{\scm}}}, U_i)$ for an arbitrary realization $\vect{x}_{\vect{A}_2}$. Moreover, if we consider $\vect{B} \coloneqq \vect{A}_1 \cup \vect{A}_2$, we have that for any arbitrary realization $\vect{x}_\vect{B}$, $f_i(\vect{x}_{\vect{B}}, \vect{X}_{\vect{PA}_i^{\graph} \setminus \vect{B}}, U_i) = f_i( \vect{X}_{\vect{PA}_i^{\graph_{\scm}}}, U_i)$. 
The same logic applies to every vertex, and removing the edges in $\vect{B}$ yields a smaller graph. Furthermore, we define SCM $\scm_{12}$ where its causal graph is obtained by taking the intersection of edges in $\graph_{\scm_1}$ and $\graph_{\scm_2}$ (alternatively, removing $\vect{B}$ from the parents of every vertex) and the link functions are obtained by applying the currying operator $f_i\big|_{\vect{X}_\vect{B} = \vect{x}_\vect{B}}$ for $\vect{B}$ defined as above for any given $i$. Hence, $\scm_{12}$ would be Markov and $P_{\scm_{12}}(\cdot) = P_\scm(\cdot)$. The only scenario where this is possible is $\graph_{\scm_1} = \graph_{\scm_{12}} = \graph_{\scm_2}$; otherwise, either $\graph_{\scm_1}$ or $\graph_{\scm_2}$ would be non-minimal since $\graph_{\scm_{12}} \subseteq \graph_{\scm_1}$ and $\graph_{\scm_{12}} \subseteq \graph_{\scm_2}$. This case, however, contradicts the fact that $\graph_{\scm_1}$ and $\graph_{\scm_2}$ were chosen differently to begin with.
\end{proof}

\begin{corollary}
Consider an SCM family compliant with the conditions of Lemma \ref{lemma:equivalence_non_constant}. In this case, there exists a single graph $\graph_0$ where the set of all topological orderings of $\graph_0$ contains all the valid causal orderings of $P_{\vect{X}}(.)$.
\end{corollary}
\begin{proof}
We only need to define $\graph_0$ as the unique graph obtained from Lemma \ref{lemma:equivalence_non_constant}. If a valid SCM $\bar{\scm}$ exists such that its ordering $\pi$ is not a valid topological ordering for $\graph_0$, then this means that the corresponding causal graph $\graph_0 \not\subseteq \bar{\graph}$, violating the minimality of $\graph_0$.
\end{proof}

\subsection{Benefits of Order Discovery versus Structure Discovery} \label{appx:order_identifiability}

In the main paper, we emphasize the value of identifying causal order by discussing that it is sufficient for intervention modelling. As elaborated in Appendix \ref{appx:notation}, we can even bypass the need for the lenient causal minimality assumption by concentrating solely on order. However, the realm of order discovery alone remains a relatively unexplored avenue in research. Even contemporary order-based methodologies such as those by \citet{cundy2021bcd} and \citet{charpentier2022differentiable}, despite having certain parameters for learning the causal ordering, concurrently introduce a large set of additional parameters to learn the full causal structure and its underlying skeleton.
Our study primarily investigates causal ordering, relying on basic algorithms like PC-KCI and CAM to identify the causal skeleton. Yet, we can achieve state-of-the-art SHD across real-world datasets. This suggests that the primary challenge in structure discovery might be determining the causal order of variables rather than estimating the graph skeleton. Thus, enhancing the order-learning phase of order-based structure discovery algorithms could be pivotal.
Concentrating on causal ordering makes the search problem more tractable, paving the way for insights from permutation learning and potentially enhancing the scalability of structure learning algorithms. In addition, performing order discovery alone facilitates the use of autoregressive networks within our ANFs. In our paper, for example, if we aimed to learn the entire structure simultaneously, we would require restrictive inductive biases in our ANFs to adhere to the graphical structure. Such constraints could hinder the network's expressiveness and compromise the optimization process of the algorithm.

In addition, we argue that in many instances, determining the order of a causal structure is the primary computational challenge (or the \emph{computational bottleneck} from an algorithmic perspective). For instance, a typical conditional independence testing algorithm used to identify the Markovian structure of a Bayesian network (such as PC) can conduct a potentially exponential number of conditional independence tests on subsets of covariates in the worst-case scenario. However, if we already know the true causal ordering, the number of tests drops to polynomial time w.r.t. $d$, offering a substantial time complexity reduction.

To be more concrete, consider having access to an oracle for conditional independence testing. If we know that covariates adhere to the true ordering $\pi = \langle \pi_1, \hdots, \pi_d \rangle$, then for any pair of nodes $(\pi_i, \pi_j)$ where $(i < j)$, the edge $(\pi_i \to \pi_j)$ exists iff the following conditional independence does not hold: $X_{\pi_i} \indep X_{\pi_j} \mid X_{\vect{\Pi}_{<j}(\pi) \setminus \{\pi_i\}}$, where $\vect{\Pi}_{<j}(\pi)$ is the set $\{\pi_1, \hdots \pi_{j-1}\}$.
Thus, by conducting $\binom{d}{2}$ conditional independence tests, we can efficiently compute the full causal structure using $\mathcal{O}(d^2)$ calls of the oracle, in contrast to the exponential calls a method like the PC algorithm might require.

\subsection{Bivariate Likelihood Ratio for the Multivariate Setting}
\label{appx:bivar-multivar}
One na\"ive approach to extend to the multivariate setting would involve using bivariate likelihood-ratio testing to find the causal direction for all pairs of nodes in the data and subsequently sorting the nodes based on these pairwise comparisons. \citet{khemakhem2021causal} suggests using a traditional constraint-based method, such as the PC algorithm, to first estimate the skeleton of the causal DAG and then orient the edges using their bivariate ratio test. However, these approaches do not apply to the class of multivariate LSNMs. In particular, we give a simple example to see why the relationship between pairs of parent-child nodes in multivariate LSNMs does not necessarily follow a bivariate LSNM with the same exogenous noises, rendering the bivariate likelihood ratio test unable to identify the correct direction. 

Consider the following SCM as the data-generating process:
\begin{align}
    &X_1 = U_1\nonumber  \\ 
    & X_2 = {X_1 } U_2  \nonumber\\
    & X_3 = \exp(X_1) + \log(X_2) + U_3
    \label{eq:scm-example}
\end{align}
where {$U_1, U_2 \sim \text{Uniform}(0, 1), U_3 \sim \mathcal{N}(0, 1)$}. \cref{eq:scm-example} follows an LSNM. Now, let's re-write $X_3$ using $X_2={X_1 }U_2$:
\begin{align}
    X_3 &= \exp(X_1) + \log(X_1) + \left(\log(U_2) + U_3\right) \nonumber \\
    X_3 &= \exp(X_1) + \log(X_1) + U'_3
    \label{eq:new-scm}
\end{align}
for $U'_3 = \log(U_2) + U_3$. However, $U'_3$ is not a normal distribution anymore. Therefore, \cref{eq:new-scm} does not follow an LSNM with a normal distribution, thus making the bivariate LSNM inapplicable.

%% file: appendices/proofs.tex
\section{LSNM Proofs}
\label{appx:proof-ordering-loss}
As discussed in the main paper, the identifiability of location-scale noise models (LSNMs) has already been proved in recent lines of work~\citep{khemakhem2021causal,immer2022identifiability,strobl2022identifying}. In particular, they define a set of extra conditions on the LSNM family, resulting in a narrower class of models called restricted LSNM, and prove identifiability for that class. Those theoretical results, however, do not distinguish between the data complexity class and the search model class. By data complexity class, we mean the complexity of the true data-generating process, while the search model class refers to the class of parametric likelihood models we use to find the correct ordering. \citet{khemakhem2021causal} showed that the model class of \textit{affine} ANFs is equivalent to the data complexity class of LSNMs. Therefore, we may want to use \method\ with affine ANFs to find the correct ordering for datasets from restricted LSNMs. However, affine ANFs do not generally comply with restricted LSNMs. Due to this mismatch between the model and data classes, we cannot apply the aforementioned theoretical results to provide the correctness of likelihood-based models such as \method. In this section, we first prove a stronger identifiability result than the previous work by \underline{not} limiting the \textit{search model class} to restricted LSNMs. We then prove the correctness of \method\ for the datasets from restricted LSNMs, where we can use any conventional affine ANF as the parametric likelihood model. Finally, we finish the section by providing proof of the equivalence of causal minimality and non-constant link functions in LSNMs, complementing our discussion in \cref{appx:assumptions}.
\subsection{Strong identifiability}
\label{appx:proof-multi}
This section provides theoretical results for the strong identifiability of LSNMs, where the search model class is an LSNM while the data complexity class is a \textit{restricted} LSNM. Our proof largely follows similar techniques used in previous work~\citep{peters2014causal,khemakhem2021causal,immer2022identifiability,strobl2022identifying}. We begin this part by defining LSNMs and restricted LSNMs.
\begin{definition}(LSNMs) An SCM $\scm$ belongs to the LSNM family if it has the following form \label{def:lsnm}
\begin{align}
    X_i = \locfunc_i(\vect{X}_{\vect{PA}^\graph_i}) + \scalefunc_i(\vect{X}_{\vect{PA}^\graph_i}) \cdot U_i, \label{eq:lsnm-simple}
\end{align}
where $\locfunc_i, \scalefunc_i: \R^{|\vect{PA}^\graph_i|} \to \R$ (positive scaling $\scalefunc_i > 0$) are continuous functions that characterize the link functions.
\end{definition}
To define the class of restricted LSNMs, we first need to define the concept of bivariate identifiability as follows:
\begin{definition}[Bivariate Identifiability]
\label{def:bi-id}
    Consider a tuple $\left(\locfunc, \scalefunc, P_X, P_U\right)$ with continuous functions $\locfunc$ and $\scalefunc > 0$ on $\R$, and independent random variables $U$ and $X$. Define 
    $$Y = \locfunc(X) + \scalefunc(X) \cdot U \ .$$
    We call the tuple \textit{bivariate identifiable} if there are \textbf{no} continuous functions $\bar{\locfunc}$ and $\bar{\scalefunc}$, s.t. the backward model holds:
    $$X = \bar{\locfunc}(Y) + \bar{\scalefunc}(Y) \cdot V\  \text{ for some $V \indep Y$ with $P_V = P_U$.}$$
\end{definition}
The standard approach to prove the identifiability in the bivariate case is to compare the data distribution from both causal and anti-causal directions and identify the constraints on the link functions such that both directions yield the same distribution. Any functional form outside these conditions will result in \textit{identifiable} families. In particular, \cite{immer2022identifiability} propose a sufficient yet general condition for bivariate identifiability by providing a partial differential equation over the link functions and showing that any function \textit{non-compliant} with the differential equation ensures an identifiable model. We now define the class of restricted LSNMs~\citep{peters2014causal,strobl2022identifying}:

\begin{condition}[Restricted LSNMs] \label{def:restricted-multi}
    Consider an LSNM (Definition~\ref{def:lsnm}). We call this model a restricted LSNM if all of its exogenous noise variables are i.i.d., and for all $i \in V(\graph)$, all $j \in \vect{PA}^\graph_i$, and all sets $\vect{C}$ with $\vect{PA}^\graph_i \setminus \{j\} \subseteq \vect{C} \subseteq  \vect{ND}^\graph_{i}\setminus \{i, j\}$ ($\vect{ND}^\graph_{i}$ are the non-descendants of $i$ in graph $\graph$), there is at least one setting of random variables in $\vect{X}_\vect{C}$ (denoted as $\vect{x}_\vect{c}$) with $p_{X_\vect{C}}(\vect{x}_\vect{c}) > 0$, s.t. the tuple
    \begin{align*}
    \big(\locfunc_i\big|_{\vect{X}_\vect{C} = \vect{x}_\vect{c}}, \scalefunc_i\big|_{\vect{X}_\vect{C} = \vect{x}_\vect{c}}, p_{X_j}(\cdot \mid \vect{X}_\vect{C} = \vect{x}_\vect{c}), p_{U_i}\big)
    \end{align*}
    satisfies bivariate identifiability in Definition~\ref{def:bi-id}. 
\end{condition}
$f\big|_{\vect{X}_\vect{C} = \vect{x}_\vect{c}}$ is the \textit{currying operator} where the function $f$ is reduced to a lower dimensional mapping by fixing a subset of input arguments $\vect{X}_\vect{C}$ to $\vect{x}_\vect{c}$. 

Using the above conditions, we are now ready to prove the strong identifiability of restricted LSNMs. To do so, we state the following two lemmas from ~\citet{peters2014causal} and refer the reader to the original paper for the proofs.

The first lemma is the following result that holds for all graphical models with the causal minimality assumptions and is first stated by \cite{peters2014identifiability} in the context of SCMs with equal error variances.

\begin{lemma}[\textbf{Proposition 29.~\citet{peters2014causal}}]
\label{lemma:multi-two-var}
Consider two SCMs $\scm$ and $\bar{\scm}$ defined over the same set of variables $\vect{X}$, where $P_{\scm} = P_{\bar{\scm}}$ but their corresponding graphs $\graph_\scm$ and $\graph_{\bar{\scm}}$ are not equal. 
Assume causal minimality holds for $\scm$ and $\bar{\scm}$. 
Then, there are $X_i, X_j \in \vect{X}$ s.t. for the sets $\vect{Q} := \vect{PA}_i^{\graph_\scm} \backslash \{j\}$, $\vect{R} := \vect{PA}_j^{\graph_{\bar{\scm}}} \backslash \{i\}$ and $\vect{C} := \vect{Q} \cup \vect{R}$ we have:
\begin{itemize}
    \item $j \to i \in E(\graph_{\scm})$ and $i \to j \in E(\graph_{\scm'})$.
    \item $\vect{C} \subseteq \vect{ND}^{\graph_\scm}_{i} \backslash \{j\}$ and $\vect{C} \subseteq \vect{ND}^{\graph_{\scm'}}_{j} \backslash \{i\}$.
\end{itemize}
\end{lemma}

The second result is a general lemma that holds for any continuous joint probability density in the presence of conditional independence. For random variables $A$ and $B$, we use $A \big|_{B=b}$ to denote the random variable $A$ after conditioning on
$B = b$, assuming density functions exist and $B$ has a positive density at $b$.

\begin{lemma}[\textbf{Lemma 36.~\citet{peters2014causal}}]
\label{lemma:technical}
    Let $Y \in \mathcal{Y}, U \in \mathcal{U}, \vect{Q}\in \mathcal{Q}, \vect{R}\in \mathcal{R}$ be random variables ($\vect{Q}$ and $\vect{R}$ can be multivariate) whose joint distribution is absolutely continuous w.r.t. some product measure with density $p_{Y, \vect{Q}, \vect{R}, U}$. Let $f: \mathcal{Y} \times \mathcal{Q} \times \mathcal{U} \to \R$ be a measurable function. If $U \indep (Y, \vect{Q}, \vect{R})$, then for all $\vect{q} \in \mathcal{Q}, \vect{r} \in \mathcal{R}$ with $p_{\vect{Q}, \vect{R}}(\vect{q}, \vect{r}) > 0$:
    $$f(Y, \vect{Q}, U)\big|_{\vect{Q} = \vect{q}, \vect{R} = \vect{r}} = f(Y\big|_{\vect{Q} = \vect{q}, \vect{R} = \vect{r}}, \vect{q}, U).$$
\end{lemma}

Given the above lemmas, we can prove the following result:
\begin{theorem}[Strong Identifiability of Restricted LSNMs]
\label{thm:structure-id}
    Assume the data $P_\vect{X}$ is generated from a restricted LSNM $\scm$ with strictly positive density w.r.t the Lebesgue measure with causal graph $\graph_\scm$. Assume $\scm$ satisfies causal minimality. Then, there is no other casual minimal LSNM $\bar{\scm}$ (\textbf{restricted or not}) with positive scaling functions $\scalefunc_i$ that has the same distribution of exogenous noises and entail the same observational distribution $P_\vect{X}$ but has a different graph $\graph_{\bar{\scm}} \neq \graph_{\scm}$.
\end{theorem}
\begin{proof}
We prove the theorem by contradiction. Assume there exist a restricted LSNMs $\scm$ and an LSNM $\bar{\scm}$, with causal minimality, such that both induce $P_\vect{X}$ while $\graph_\scm \neq \graph_{\bar{\scm}}$. Then, let $X_i$ and $X_j$ be the two random variables that follow Lemma~\ref{lemma:multi-two-var}. In particular, since $X_j$ is a parent of $X_i$ in $\graph_\scm$ and $X_i$ is a parent of $X_j$ in $\graph_{\bar{\scm}}$, then, we have the following:
\begin{align}
    &X_i = \locfunc_i(\vect{X}_{\vect{PA}^{\graph_\scm}_i \backslash \{j\}}, X_j) + \scalefunc_i(\vect{X}_{\vect{PA}^{\graph_\scm}_i \backslash \{j\}}, X_j)\cdot U_i,\quad U_i \indep  \vect{X}_{\vect{PA}^{\graph_\scm}_i} \label{eq:helper-L}\\[0.5em]
    &X_j = \bar{\locfunc}_j(\vect{X}_{\vect{PA}^{\graph_{\bar{\scm}}}_j \backslash \{i\}}, X_i) + \bar{\scalefunc}_j(\vect{X}_{\vect{PA}^{\graph_{\bar{\scm}}}_j \backslash \{i\}}, X_i)\cdot U_j,\quad U_j \indep  \vect{X}_{\vect{PA}^{\graph_{\bar{\scm}}}_j} \label{eq:helper-Y}
\end{align}

Moreover, let $\vect{C} =  \vect{PA}_i^{\graph_\scm}  \backslash \{j\} \bigcup \vect{PA}_j^{\graph_{\bar{\scm}}}  \backslash \{i\}$ be the corresponding set defined in Lemma~\ref{lemma:multi-two-var}. In particular, define $X_i^* \coloneqq X_i|_{\vect{X}_\vect{C} = \vect{x}_\vect{C}}$ and $X_j^* \coloneqq X_j|_{\vect{X}_\vect{C} = \vect{x}_\vect{C}}$ for an arbitrary value of $\vect{x}_{\vect{C}}$ with $P(\vect{x}_{\vect{C}}) > 0$. $\vect{C}$ is a subset of non-descendents of $i$ and $j$, and therefore, exogenous noises $U_i$ of $\scm$ and ${U}_j$ of $\bar{\scm}$ are both independent from $\vect{X}_\vect{C}$. 

Now, consider the continuous (and measurable) link functions $\locfunc_i$ and $\scalefunc_i$ in $\scm$. We can use Lemma \ref{lemma:technical} by setting $Y = X_j$, $\vect{Q} = \vect{X}_{\vect{PA}^{\graph_\scm}_i \setminus \{j\}}$, $\vect{R} = \vect{X}_{\vect{PA}^{\graph_{\bar{\scm}}}_j \setminus \{i\}}$, and conditioning on $\vect{X}_\vect{C} = \vect{x}_\vect{C}$, which is the same as conditioning over both $\vect{R} = \vect{r}$ and $\vect{Q} = \vect{q}$. Therefore,
\begin{equation}\label{eq:helper-lstar}
    X_i^* = \locfunc_i\big|_{\vect{X}_\vect{C}=\vect{x}_\vect{C}}(X_j^*) + \scalefunc_i\big|_{\vect{X}_\vect{C}=\vect{x}_\vect{C}}(X_j^*)\cdot U_i,\quad U_i \indep  X_j^*.
\end{equation}

Similarly, since $P_{\bar{\scm}} = P_\scm$ and the link functions are continuous in $\bar{\scm}$, we can show that:
\begin{equation}\label{eq:helper-ystar}
    X_j^* = \bar{\locfunc}_j\big|_{\vect{X}_\vect{C}=\vect{x}_\vect{C}}(X_i^*) + \bar{\scalefunc}_j\big|_{\vect{X}_\vect{C}=\vect{x}_\vect{C}}(X_i^*)\cdot {U}_j,\quad {U}_j \indep  X_i^*.
\end{equation}

$\scm$ follows a restricted LSNM as defined by Condition \ref{def:restricted-multi} and set $\vect{C}$ follows the condition  $\vect{PA}_i^{\graph_\scm} \setminus \{j\} \subseteq \vect{C} \subseteq \vect{ND}_i^{\graph_\scm} \setminus \{i, j\}$. Therefore, the tuple 
$$\left(\locfunc_i\big|_{\vect{X}_\vect{C} = \vect{x}_\vect{C}}, \scalefunc_i\big|_{\vect{X}_\vect{C} = \vect{x}_\vect{C}}, P_{X^*_j}, P_{U_i}\right)$$
must satisfy bivariate identifiability for at least one realizable setting of $\vect{X}_\vect{C} = \vect{x}_\vect{C}$. However, if such $\bar{\scm}$ exist, \eqref{eq:helper-ystar} implies that a backward model can be constructed for every realizable setting of $\vect{X}_\vect{C}$ since ${U}_j$ and ${U}_i$ have the same distribution (i.i.d. assumption of exogenous noises in restricted LSNM) and the model specification of $\bar{\scm}$ complies with the backward model in Definition \ref{def:bi-id}. Hence, the contradiction yields our proof.
\end{proof}

It is important to note that Theorem \ref{thm:structure-id} presents a more general result than Theorem 3 in \citet{strobl2022identifying}. It shows that, for any restricted LSNM, there is no other restricted LSNM \textit{and no other {non-restricted} LSNM} with the same observational distribution and causal graph. As we will see in the next section, this result is crucial for demonstrating the convergence of \method\ to the correct causal ordering. In fact, affine ANFs might not always satisfy restricted LSNMs. However, since they always produce LSNMs with continuous link functions and positive scaling, we can use Theorem~\ref{thm:structure-id} to prove their convergence.

\subsection{Affine ANFs for Causal Discovery}
\label{appx:affine-anf-disc}
In this section, we prove that if the data is generated from a restricted LSNM, in the limit of infinite data $(N \to \infty)$, only a valid causal ordering maximizes the parametric likelihood of a conventional affine ANF. Thus, with sufficient data and an effective optimization method, the MLE framework of \method~reliably extracts a valid causal ordering. Before continuing, we formally define a conventional affine ANF w.r.t. a restricted LSNM:

\begin{definition}[Conventional Affine ANF] \label{def:affine-anf}
    An affine ANF parameterized by $\theta \in \Theta$ maps a set of latent noise variables to data distribution using an affine transform with location functions $\locfunc_i^{\theta}$ and scale functions $\scalefunc_i^\theta$. We call such affine ANF \textit{conventional} w.r.t. a given restricted LSNM $\scm$, if it has the following properties:
    \begin{enumerate}[leftmargin=*]
        \item The modelled location and scale functions $\locfunc_i^\theta$ and $\scalefunc_i^\theta$ are continuous.
        \item There is a parameter setting $\theta \in \Theta$ such that, for any possible location and scale function $\locfunc_i$ and $\scalefunc_i$ in $\scm$, we have $\locfunc_i^\theta(\cdot) = \locfunc_i(\cdot)$ and $\scalefunc_i^\theta(\cdot) = \scalefunc_i(\cdot)$.
        \item The flow maps from the latent space of noise variables $P_\vect{U}$ to the data space where $U_1, \ldots, U_d \sim P_{\vect{U}}$ are i.i.d and follow that same distribution as the exogenous noise in $\scm$.
        \item The scaling functions $\scalefunc_i^\theta$ are positive.
        \item The entailed distribution from the ANF has strictly positive density w.r.t. the Lebesgue measure.
    \end{enumerate}
\end{definition}

All these properties are easy to satisfy in practice using inductive biases in the network: (1) is satisfied if the network activations are continuous; (2) is satisfied if the neural networks modelling the scale and noise functions are sufficiently large; (3) is satisfied when all the latent noise variables are set to a single distribution family respecting the restricted LSNM; (4) also holds when the output of the scaling function is passed through exponentiation; and finally, (5) is satisfied when the latent noise distribution also has strictly positive density w.r.t the Lebesgue measure like Gaussian noise. Although these conditions might not be strictly required in practice, defining them allows us to develop a theoretical basis to prove the soundness of \method.

Such ANFs can be parameterized using a single ensemble through masking, as detailed in \cref{sec:method} and \cref{appx:arch}. Notably, \cite{khemakhem2021causal} demonstrated that affine transforms are transitive and stacking multiple affine transforms can still indirectly model $\locfunc_i^\theta$ and $\scalefunc_i^\theta$. Thus, rather than explicitly increasing the expressiveness of the neural network architectures for $\locfunc_i^\theta$ and $\scalefunc_i^\theta$ by widening the networks or making them deeper, one can also stack multiple transforms to enhance expressiveness implicitly. This might potentially reduce the number of parameters needed for modelling.

We now prove that the affine ANFs that maximize the likelihood will result in a valid ordering for a restricted LSNM.

\anfcorrectprop*
\begin{proof}
We first consider the KL-divergence between $P_{\scm}(\cdot)$ and $P_{\vect{X}}(\cdot; \pi, \theta)$:

\begin{align}
    \text{KL}(P_{\scm}(\cdot) \| P_{\vect{X}}(\cdot; \pi, \theta)) =  \E_{\vect{x} \sim P_{\scm}}\left[\log P_{\scm}(\vect{x})\right] - \E_{\vect{x} \sim P_{\scm}}\left[\log P_{\vect{X}}(\vect{x}; \pi, \theta)\right] \geq 0
\end{align}
Therefore,
\begin{align}
     \ell^*_{\infty}(\pi) = \min_{\theta} \E_{\vect{x} \sim P_{\scm}}\left[-\log P_{\vect{X}}(\vect{x}; \pi, \theta)\right] \geq \E_{\vect{x} \sim P_{\scm}}\left[-\log P_{\scm}(\vect{x})\right],
\end{align}
where the equality holds iff $P_{\vect{X}}(\cdot; \pi, \theta) = P_{\scm}(\cdot)$ almost everywhere. Therefore, the parameters $\pi$ and $\theta$ that minimize $\ell_\infty^*$ are the ones that yield the same distribution as $P_\scm$. Hence, we need to prove the following statements:
\begin{enumerate}
    \item For each $\pi \in \Pi_{\graph_\scm}$, there exists a $\theta_\pi$ such that $P_\scm(\cdot) = P_{\vect{X}}(\cdot; \pi, \theta_\pi)$ almost everywhere, and
    \item  $P_\scm(\cdot)$ is not equal to $P_{\vect{X}}(\cdot; \bar{\pi}, \bar{\theta})$ for all $\bar{\pi} \not\in \Pi_{\graph_\scm}$ and all $\bar{\theta} \in \Theta$.
\end{enumerate} 
\paragraph{Statement (1)} Consider the location and scale functions of $\scm$ itself. For a given valid ordering $\pi$, we know that $\vect{PA}_{\pi(i)}^{\graph_\scm} \subseteq \vect{\Pi}_{< i}(\pi) \coloneqq \{\pi(1), \hdots, \pi(i-1)\}$. Therefore, we can add arguments to the location and scale functions without changing their output. The rest follows immediately from property (2) in Definition~\ref{def:affine-anf}.

\paragraph{Statement (2)} Consider an arbitrary $\bar{\pi} \notin \Pi_{\graph_\scm}$ and define $\graph^{(0)}_{\bar{\pi}}$ as the fully-connected causal graph with ordering $\bar{\pi}$. Moreover, consider an arbitrary conventional affine ANF $\bar{\theta} \in \Theta$. We denote $\scm^{(0)}$ as the LSNM corresponding to $\bar{\theta}$ and define the following set:
\begin{align}
    E^{(0)} := \{(j \to i) \in E(\graph^{(0)}_{\bar{\pi}});\ \scalefunc^{\bar{\theta}}_{i} \text{ and } \locfunc^{\bar{\theta}}_{i} \text{ are constant w.r.t. } X_j\}.
\end{align}
If $E^{(0)}$ is an empty set, then the equivalence of causal minimality and non-constant link functions (Appendix~\ref{appx:LSNM-minimality}) implies that $\scm^{(0)}$ is causal-minimal. Otherwise, define $\scm^{(1)}$ with the same noise distribution as $\scm^{(0)}$, the causal graph $\graph^{(1)}_{\bar{\pi}} = \graph^{(0)}_{\bar{\pi}} \setminus E^{(0)}$, and link functions $\bar{\locfunc}^{(1)}_i := \locfunc^{\bar{\theta}}_i \big|_{\vect{X}_{\vect{B}^{(0)}_i} = \vect{x}_{\vect{B}^{(0)}_i}}$, $\bar{\scalefunc}^{(1)}_i := \scalefunc^{\bar{\theta}}_i \big|_{\vect{X}_{\vect{B}^{(0)}_i} = \vect{x}_{\vect{B}^{(0)}_i}}$, where $\vect{B}^{(0)}_i := \{j \in V(\graph^{(0)}_{\bar{\pi}}); \ (j \to i) \in E^{(0)}\}$. Note that
 $P_{\scm^{(1)}}=P_{\scm^{(0)}}$. Now, we can use a similar procedure to define $E^{(1)}, \scm^{(2)}, \ldots$ until we get to a causal-minimal LSNM $\bar{\scm}$ with $\graph^{\bar{\scm}}_{\bar{\pi}} \subseteq \graph^{(0)}_{\bar{\pi}}$ and $P_{\bar{\scm}} = P_{\scm^{(0)}} = P_{\vect{X}}(\cdot; \bar{\pi}, \bar{\theta})$. However, $\graph_{\bar{\pi}}^{\bar{\scm}} \neq \graph_{\scm}$ since $\graph_{\scm}$ is not a sub-graph of $\graph^{(0)}_{\bar{\pi}}$ ($\bar{\pi}$ is not a valid ordering of $\graph_{\scm}$). Theorem \ref{thm:structure-id} concludes that $P_{\bar{\scm}} \neq P_{\scm}$, proving our result.
\end{proof}

\subsection{Equivalence of Causal Minimality and Non-Constant Link Functions in LSNMs} \label{appx:LSNM-minimality}

In Appendix \ref{appx:notation}, we discussed causal minimality and non-constant SCMs. Here, we first show that causal minimality implies non-constant link functions for general SCMs in Lemma~\ref{lemma:one-side-minimal}. We then prove that the two concepts are equivalent for the family of LSNMs in Lemma~\ref{lemma:two-side-minimal}.

\begin{lemma} 
\label{lemma:one-side-minimal}
Suppose an SCM $\scm = (\graph, \Fcal, P_{\vect{U}})$ is causal-minimal. Then, $\scm$ is a non-constant SCM. That is, for each $i \in V(\graph)$ and $j \in \vect{PA}^\graph_i$, there are  $u_i \sim P_{U_i}$ and realizations $(\vect{x}_{\vect{PA}^\graph_i \backslash{\{j\}}}, x_j)$ and $(\vect{x}_{\vect{PA}^\graph_i \backslash{\{j\}}}, x'_j)$ such that $x_j \neq x'_j$ and
    $$f_i(\vect{X}_{\vect{PA}^\graph_i \backslash{\{j\}}} = \vect{x}_{\vect{PA}^\graph_i \backslash{\{j\}}}, X_j = x_j, U_i=u_i) \neq f_{i}(\vect{X}_{\vect{PA}^\graph_i \backslash{\{j\}}} =\vect{x}_{\vect{PA}^\graph_i \backslash{\{j\}}}, X_j=x'_j, U_i=u_i).$$
\end{lemma}
\begin{proof}
We prove the result by contradiction. 
Suppose there exists $i \in V(\graph)$ and $j \in \vect{PA}^\graph_i$ such that for all realizations $\vect{x}_{\vect{PA}^\graph_i \backslash{\{j\}}}$, $u_i$, and $x_j \neq x'_j$, we have
    \begin{align}f_i(\vect{X}_{\vect{PA}^\graph_i \backslash{\{j\}}} = \vect{x}_{\vect{PA}^\graph_i \backslash{\{j\}}}, X_j = x_j, U_i=u_i) = f_{i}(\vect{X}_{\vect{PA}^\graph_i \backslash{\{j\}}} =\vect{x}_{\vect{PA}^\graph_i \backslash{\{j\}}}, X_j=x'_j, U_i=u_i).
    \label{eq:non-const-mech}
    \end{align}
    Consider the SCM $\bar{\scm}=(\bar{\graph}, \bar{\mathcal{F}}, P_{\vect{U}})$, where $\bar{\graph} = \graph \setminus \{j\to i\}$, and each $\bar{f}_i \in \bar{\mathcal{F}}$ is defined as 
    \begin{align}
\bar{f}_i(\vect{X}_{\vect{PA}^{\bar{\graph}}_i}= \vect{x}, U_i = u_i) \coloneqq f_i(\vect{X}_{\vect{PA}^{\graph}_i \backslash{\{j\}}} = \vect{x}, X_j = 0, U_i = u_i). 
    \end{align}
    Hence, \eqref{eq:non-const-mech} implies $P_{\bar{\scm}} = P_{\scm}$. However, $\bar{\graph} \subsetneq \graph$, which contradicts the causal minimality of $\scm$.
\end{proof}

\begin{lemma}
\label{lemma:two-side-minimal}
Suppose an LSNM $\scm = (\graph, \Fcal, P_{\vect{U}})$ is non-constant, i.e., for each $i \in V(\graph)$ and $j \in \vect{PA}^\graph_i$, there is a value $u_i \sim P_{U_i}$ and realizations $(\vect{x}_{\vect{PA}^\graph_i \backslash{\{j\}}}, x_j)$ and $(\vect{x}_{\vect{PA}^\graph_i \backslash{\{j\}}}, x'_j)$ such that $x_j \neq x'_j$ and
\begin{align}
    \locfunc_i(\vect{x}_{\vect{PA}^\graph_i \backslash{\{j\}}}, x_j) + \scalefunc_i(\vect{x}_{\vect{PA}^\graph_i \backslash{\{j\}}}, x_j) \cdot u_i \neq
    \locfunc_i(\vect{x}_{\vect{PA}^\graph_i \backslash{\{j\}}}, x'_j) + \scalefunc_i(\vect{x}_{\vect{PA}^\graph_i \backslash{\{j\}}}, x'_j) \cdot u_i.\label{eq:technical-inequality}
\end{align}
Then, $\scm$ is causal-minimal.
\end{lemma}
\begin{proof}
Without loss of generality, we assume that the LSNM family has exogenous noises $P_{U_i}$ with zero mean and unit variance. We prove the result by contradiction. Assume $\scm$ is not causal-minimal. Then, there is an SCM $\bar{\scm}$ with causal graph $\graph_{\bar{\scm}} \subsetneq \graph_\scm$ such that $P_{\bar{\scm}} = P_\scm$. This means there exists a node $i$ such that $\vect{PA}_i^{\graph_{\bar{\scm}}} \subsetneq \vect{PA}_i^{\graph_\scm}$. Choose a $j \in \vect{PA}_i^{\graph_\scm} \setminus \vect{PA}_i^{\graph_{\bar{\scm}}}$. Since $j \to i \notin E(\graph_{\bar{\scm}})$, we have $X_i \indep X_j \mid \vect{X}_{\vect{PA}^\graph_i \backslash{\{j\}}}$. Therefore for any realizations $(\vect{x}_{\vect{PA}^\graph_i \backslash{\{j\}}}, x_j)$ and $(\vect{x}_{\vect{PA}^\graph_i \backslash{\{j\}}}, x'_j)$, we must have 
\begin{align}
&\E\left[X_i \mid \vect{X}_{\vect{PA}^\graph_i \backslash{\{j\}}} = \vect{x}_{\vect{PA}^\graph_i \backslash{\{j\}}}, X_j = x_j \right] = \E\left[X_i \mid \vect{X}_{\vect{PA}^\graph_i \backslash{\{j\}}} = \vect{x}_{\vect{PA}^\graph_i \backslash{\{j\}}}, X_j = x'_j \right] \label{eq:first-moment-eq}\\
&\Var\left[X_i \mid \vect{X}_{\vect{PA}^\graph_i \backslash{\{j\}}} = \vect{x}_{\vect{PA}^\graph_i \backslash{\{j\}}}, X_j = x_j \right] = \Var\left[X_i \mid \vect{X}_{\vect{PA}^\graph_i \backslash{\{j\}}} = \vect{x}_{\vect{PA}^\graph_i \backslash{\{j\}}}, X_j = x'_j \right] \label{eq:second-moment-eq}
\end{align}
Re-writing the L.H.S. in \eqref{eq:first-moment-eq}, we get
\begin{align*}
  &\E\left[ \locfunc_i(\vect{X}_{\vect{PA}^\graph_i \backslash{\{j\}}}, X_j) + \scalefunc_i(\vect{X}_{\vect{PA}^\graph_i \setminus \{j\}}, X_j) \cdot U_i  \mid \vect{X}_{\vect{PA}^\graph_i  \backslash{\{j\}}} = \vect{x}_{\vect{PA}^\graph_i  \backslash{\{j\}} }, X_j = x_j\right]\\
  & =\locfunc_i(\vect{x}_{\vect{PA}^\graph_i \setminus \{j\}}, x_j)  + \scalefunc_i(\vect{x}_{\vect{PA}^\graph_i \setminus \{j\} }, x_j) \cdot \E\left[U_i\right]  \tag*{(continuity of link functions in our definition of LSNM)}\\
  & = \locfunc_i(\vect{x}_{\vect{PA}^\graph_i \setminus \{j\}}, x_j) \tag*{(zero-mean exogenous noise)}
\end{align*}
Similarly, we can re-write the L.H.S. variance in \eqref{eq:second-moment-eq}:
\begin{align*}
  &\Var\left[ \locfunc_i(\vect{X}_{\vect{PA}^\graph_i \backslash{\{j\}}}, X_j) + \scalefunc_i(\vect{X}_{\vect{PA}^\graph_i \setminus \{j\}}, X_j) \cdot U_i  \mid \vect{X}_{\vect{PA}^\graph_i  \backslash{\{j\}}} = \vect{x}_{\vect{PA}^\graph_i  \backslash{\{j\}} }, X_j = x_j\right]\\
  & = \scalefunc^2_i(\vect{x}_{\vect{PA}^\graph_i \setminus \{j\} }, x_j) \cdot \Var\left[U_i\right] \\
  & = \scalefunc^2_i(\vect{x}_{\vect{PA}^\graph_i \setminus \{j\} }, x_j) \tag*{(unit-variance exogenous noise)}
\end{align*}
We can derive similar equalities for $X_j = x'_j$. Therefore, we must have  
$\locfunc_i(\vect{x}_{\vect{PA}^\graph_i \setminus \{j\}}, x_j) = \locfunc_i(\vect{x}_{\vect{PA}^\graph_i \setminus \{j\}}, x'_j)$ and
$\scalefunc_i(\vect{x}_{\vect{PA}^\graph_i \backslash{\{j\}}}, x_j) = \scalefunc_i(\vect{x}_{\vect{PA}^\graph_i \backslash{\{j\}}}, x'_j)$ (although we showed the square of scale functions are equal, the positivity of scale functions in our LSNM definition yields this), which contradicts \eqref{eq:technical-inequality}.
\end{proof}

%% file: appendices/algo.tex
 \section{Algorithm Details}
  
\subsection{Network Architecture} \label{appx:arch}
The ANF architecture used in \method\ is parameterized by a set of masked autoregressive MLPs \citep{germain2015made}. In masked MLPs, hidden neurons are assigned to a label ranging from $1$ to $d$, and they only connect to subsequent layer neurons with labels greater than or equal to their own, yielding a dependency graph where outputs are dependent on inputs with smaller labels. 
In other words, given an ordering $\pi$ for the data, neurons $i$ and $j$ of consecutive layers can connect only if $\pi(i) \le \pi(j)$. For example, consider an $l$-layer network with parameters $W^{(1)}, \hdots, W^{(l)}$, and activation function $\sigma$. The network's output is as follows:
{
\begin{equation} \label{eq:simplistic-masked-mlp} 
         \sigma\left( \hat{\vect{M}}_\pi \odot W^{(l)} \sigma \left( \vect{M}^{(l-1)}_\pi \odot W^{(l-1)}  \sigma \left(\ldots  \vect{M}^{(1)}_\pi \odot W^{(1)} \vect{u} \right)\right)\right),
\end{equation}
}%
where the masking matrices are given by $\big(\text{Label}^{(k)}_i$ denotes the label of neuron $i$ at layer $k\big)$:
{
\begin{align} 
& \label{eq:simplistic-masked-mlp2} \forall k \leq l-1:\ \vect{M}^{(k)}_\pi({i,j}) =
\mathbb{I}\left[\pi(\text{Label}^{(k)}_i) \le \pi(\text{Label}^{(k)}_j)\right], \text{and} \quad \hat{\vect{M}}_\pi({i,j}) =
\mathbb{I}\left[\pi(\text{Label}^{(l)}_i) < \pi(\text{Label}^{(l)}_j)\right].
\end{align}
}%
Matrices $\vect{M}^{(k)}_\pi$ and $\hat{\vect{M}}_\pi$  can also be written as:
{
\begin{align}
    \vect{M}^{(k)}_\pi = \vect{P}_\pi^\top \vect{M}^{(k)}_\idenperm \vect{P}_\pi, \quad 
    \hat{\vect{M}}_\pi = \vect{P}_\pi^\top \hat{\vect{M}}_\idenperm \vect{P}_\pi, 
\end{align}
}%
where $\vect{M}^{(k)}_\idenperm$ and $\hat{\vect{M}}_\idenperm$ are lower triangular binary matrices with identity diagonal entries and zero diagonals, respectively.

\begin{figure}[t]
\captionsetup{font=footnotesize,labelfont=footnotesize}
    \centering
    \includegraphics[width=0.8\textwidth]{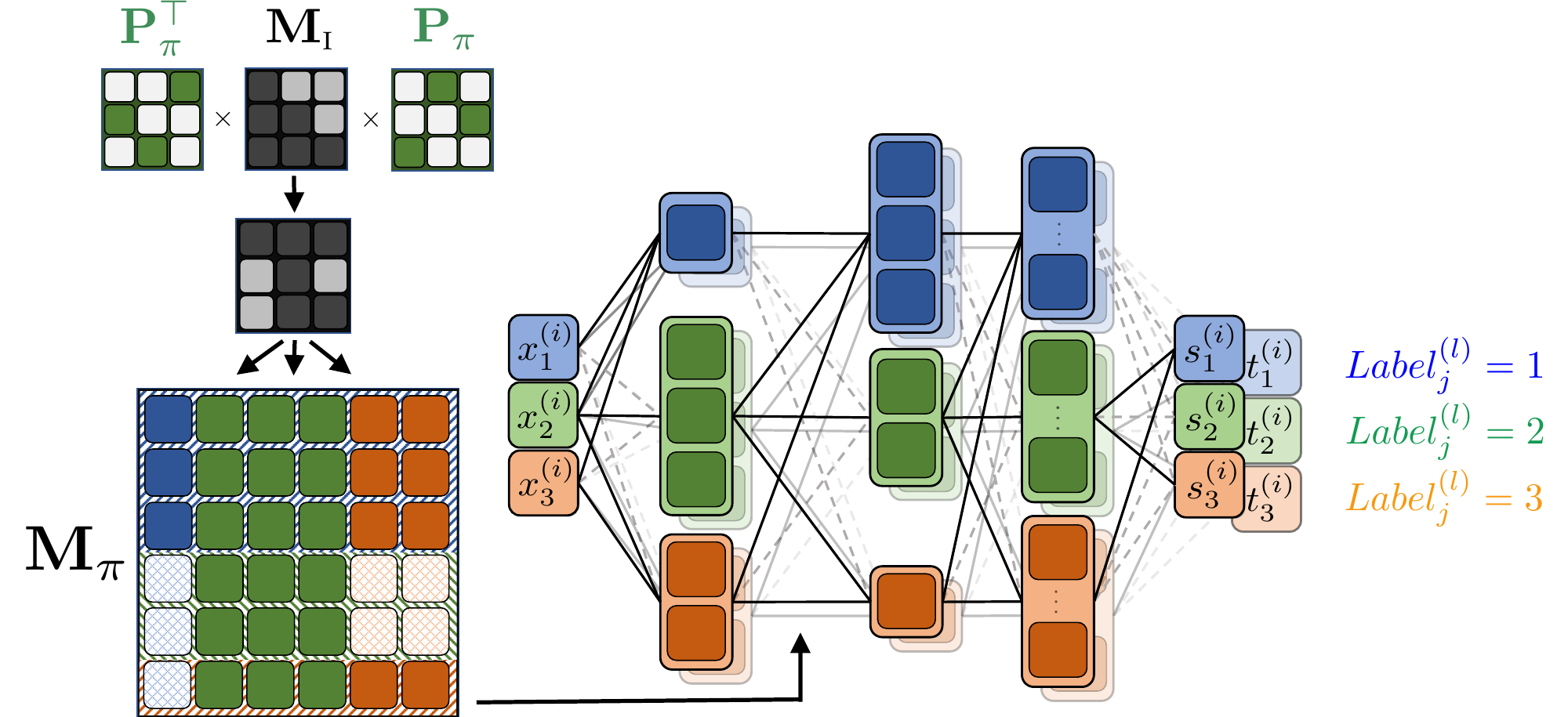}
    \caption{Network architecture of \method\ for the special case of affine ANFs with block masking matrices. The causal ordering of variables is $\langle X_2, X_3, X_1 \rangle$. Each colour represents the neurons corresponding to one variable. In the final layer, self-connections are absent, and each neuron depends solely on inputs with smaller labels.}
    \label{fig:arch-deep-dive}
\end{figure}

We allow each neuron in the masked MLP to be multi-dimensional to make the network more expressive. In other words, the dimension of each layer in equations~\ref{eq:simplistic-masked-mlp} and~\ref{eq:simplistic-masked-mlp2} can be a multiple of $d$. In this formulation, each masking will be a block matrix, and its $(i,j)$th block will have a dimension of $\dim^{(l)}_i \times \dim^{(l+1)}_j$, where $\dim^{(l)}_i$ denotes the dimension of the $i$th neuron in layer $l$. \autoref{fig:arch-deep-dive} illustrates an example of this architecture for the special case of affine ANFs.

One distinction between \method~and other similar permutation learning causal discovery algorithms such as VI-DP-DAG~\citep{charpentier2022differentiable} is that we incorporate the permutation directly into our architecture. In contrast, other works generally apply permutations to the input covariates and then pass it to the model. In the latter case, for any given input configuration following a permutation, the weight between neurons $i$ and $j$ has a different meaning. For example, when the input configuration follows an identity permutation $\idenperm$, the weight between neurons related to covariate $X_1$ and neurons related to covariate $X_2$ influence the link function $f_2$. However, assume that we now consider a different ordering $\idenperm_{12}$ which is obtained by swapping the first and second covariate in the identity permutation. 
Now the same neurons that modelled the link function of $X_2$ would model the link function of $X_1$, and therefore, the information from one input configuration does not easily transfer to another input configuration.
That being said, it can be seen that with our modelling technique, all weights consistently correspond to specific link functions, even if the permutation changes. As an example, training the network on the permutation $\idenperm$ can still yield transferable information to $\idenperm_{12}$, especially concerning the interdependencies other than the ones between $X_1$ and $X_2$. We believe that this allows for parameter sharing and increased utilization of training.

Our architecture also allows for end-to-end training where no tricks such as the straight-through gradient estimator are needed. To do so, we can leverage the fact that for a sufficiently low temperature $\tau$, the matching function $M(\StructureBelief)$ is closely approximated by the Sinkhorn function $S(\StructureBelief / \tau)$. We can thus directly use the soft samples from the Sinkhorn operator as the permutation matrix $\HardPermutation$ in the forward and backward passes, making the entire proxy score differentiable w.r.t. both $\StructureBelief$ and the ANF parameters.
We call this approach ``Soft-Sinkhorn'' and run experiments on it. However, we observe that this method learns to cheat. Specifically, since the whole optimization procedure is end-to-end, the model can potentially modify the value of $\StructureBelief$ such that its scale cancels the effect of small $\tau$ (e.g. make the parameters of $\StructureBelief$ arbitrarily small to match the scale of $\tau$). $\tau$ being relatively small is a necessary condition for the Sinkhorn function to generate valid permutation matrices, and as a result, such changes might cause the matrices generated by the Sinkhorn function to evolve into soft doubly stochastic matrices that do not necessarily align with the Matching function. Subsequently, masking matrices $\vect{M}$ in \eqref{eq:simplistic-masked-mlp} obtained from these doubly stochastic matrices can potentially introduce loops between neurons of certain covariates when considering the computation graph of our ANF. Note that the entire framework hinges on the fact that our networks are autoregressive and these loops can allow the model to cheat and adversely influence the training process. We empirically illustrate this phenomenon in \autoref{tab:soft-fails}, for a small synthetic data (Parametric path with $4$ vertices) where the proxy score function is unnaturally high; meaning that it has probably learned to cheat and obtain a likelihood above the maximum theoretically possible bound.

\begin{table}
  \centering
    \caption{\small A comparison between \method~and the Soft Sinkhorn method across our small synthetic datasets. Although a higher proxy score is preferable, Soft Sinkhorn yields misleadingly high score values while generating near-random orderings as captured by \cbc, implying self-loops.}
    \begin{adjustbox}{width=0.5\linewidth,center}
  \begin{tabular}{ccc}
    \toprule
    Method & Mean \cbc & Mean Proxy Score \\
    \midrule
    \method~(Soft-Sinkhorn) & 0.49 & \textbf{8.77} \\
    \method & \textbf{0.14} & -2.38 \\
    \method~(Gumbel-Sinkhorn) & 0.35 & -2.53 \\
    \bottomrule
  \end{tabular}
  \end{adjustbox}
  \label{tab:soft-fails}
\end{table}

\subsection{Order Learning Details}\label{appx:algo}

Before stating the algorithm, we discuss the concrete loss function used in our implementation for the special case of affine ANFs. Recall that, for a given ordering $\pi$, affine ANFs parameterized by $\Theta$ use the following transform functions to model the  data:
{
\begin{align}
    X_{\pi(j)} = \locfunc^\theta_{\pi(j)}\left(\vect{X}_{< \pi(j)}\right) + \scalefunc^\theta_{\pi(j)}\left(\vect{X}_{< \pi(j)}\right) \cdot U_{\pi(j)}, \quad \forall j \in [d],
    \label{eq:affine-anf-conc}
\end{align}
}%
where ${<\pi(j)}$ corresponds to $\pi(1), \ldots, \pi(j-1)$, and functions $\{\scalefunc_{i}^\theta\}_{i=1}^{d}$ are all positive (typically via exponentiation). Now, for a given dataset $\data = \{\vect{x}^{(i)}\}_{i=1}^N$, we can use the change-of-variables formula in \eqref{eq:change-var}, and the affine ANF data-generating process described in \eqref{eq:affine-anf-conc} to re-write the loss function defined in \eqref{eq:max-log-like} as follows:
{
\begin{align}
     \ell_N(\pi ; \theta) &= \frac{1}{N} \sum_{i=1}^N \sum_{j=1}^d \log P_{U_{\pi(j)}}\left(\frac{x^{(i)}_{\pi(j)} - t^\theta_{\pi(j)}\left(\vect{x}^{(i)}_{< \pi(j)}\right)}{s^\theta_{\pi(j)}\left(\vect{x}^{(i)}_{< \pi(j)}\right)}\right) - \log s^\theta_{\pi(j)}\left(\vect{x}^{(i)}_{< \pi(j)}\right) \\
     &= \text{const.} + \frac{1}{N} \sum_{i=1}^N \sum_{j=1}^d  \left(\frac{x^{(i)}_{\pi(j)} - t^\theta_{\pi(j)}\left(\vect{x}^{(i)}_{< \pi(j)}\right)}{s^\theta_{\pi(j)}\left(\vect{x}^{(i)}_{< \pi(j)}\right)}\right)^2 - \log s^\theta_{\pi(j)}\left(\vect{x}^{(i)}_{< \pi(j)}\right)\  \text{(standard normal base distribution)} \nonumber
\end{align}
}%
We describe \method\ in Algorithm~\ref{alg:lsnm}. The function $\textsc{GumbelNoise}$ anneals the standard deviation of the Gumbel noise based on the epoch number, ensuring convergence of the distribution over permutations. Moreover, in practice, instead of the one-step alternation scheme shown in Algorithm~\ref{alg:lsnm}, we employ a phase-changing scheduler to alternate between optimizing for $\theta$ and $\StructureBelief$ -- See \autoref{appx:experimental-details}.
 \begin{algorithm}
\caption{\method}
\label{alg:lsnm}
\begin{algorithmic}[1]
\REQUIRE Dataset $\mathcal{D} = \{\vect{x}^{(i)}\}_{i=1}^N$, Max num of hard samples $k$, Learning rates $\lambda_\theta, \lambda_\StructureBelief$, Max epoch $L$\\
\ENSURE Estimated causal ordering
\STATE Initialize $\StructureBelief^{(0)}$ and $\theta^{(0)}$
\FOR{$l = 0, 1, \ldots, L$}
    \STATE $\vect{\epsilon}_1, \ldots, \vect{\epsilon}_k \gets \textsc{GumbelNoise}(l)$ \hfill \# generate $k$ Gumbel noises 
    \STATE $\HardPermutation_1, \ldots, \HardPermutation_{k} \gets M(\StructureBelief^{(l)} + \vect{\epsilon}_1), \ldots, M(\StructureBelief^{(l)} + \vect{\epsilon}_k)$ \hfill \# generate hard permutations~\vspace{0.25em}
    \STATE $\mathcal{H}_k \gets \textsc{Unique}(\{\HardPermutation_{1}, \ldots, \HardPermutation_k\})$~\vspace{0.25em}
    \STATE $\hat{\ProxyLoss}_k(\StructureBelief, \theta) \gets {\sum_{\HardPermutation_\pi \in \mathcal{H}_k} \exp{\langle \StructureBelief, \HardPermutation_\pi \rangle_F} \cdot \ell_N(\pi; \theta)}\;/\; {\sum_{\HardPermutation_{\pi} \in \mathcal{H}_k} \exp{\langle \StructureBelief, \HardPermutation_{\pi} \rangle_F}}$ \hfill \# proxy score~\vspace{0.25em}
    \STATE $\theta^{(l+1)} \gets \theta^{(l)} + \lambda_\theta \cdot \nabla_{\theta} \hat{\ProxyLoss}_k(\StructureBelief^{(l)}, \theta^{(l)})$~\vspace{0.25em}
    \STATE $\StructureBelief^{(l+1)} \gets \StructureBelief^{(l)} + \lambda_\StructureBelief \cdot \nabla_{\StructureBelief} \hat{\ProxyLoss}_k(\StructureBelief^{(l)}, \theta^{(l+1)})$
\ENDFOR
\STATE $\hat{\HardPermutation} = \textsc{Mode}(\{\HardPermutation_{1}, \ldots, \HardPermutation_k\})$ \hfill \# choose the most frequently generated permutation
\RETURN $\textsc{Ordering}(\hat{\HardPermutation})$ \hfill \# return the corresponding topological ordering
\end{algorithmic}
\end{algorithm}

%% file: appendices/experiment_setup.tex
\section{Experimental Details} \label{appx:experimental-details}
\xhdr{Gumbel Noise Scheduling}
\method~incorporates a strategy of annealing the Gumbel noises added to $\StructureBelief$. We use a large standard deviation for the Gumbel noise during the initial phase, ensuring a broad coverage of the permutation set $\mathcal{H}_k$. As training proceeds, we move to a smaller locality by reducing the Gumbel noise standard deviation to zero. This approach ensures that generated permutations are centered on the single point $M(\StructureBelief)$, leading to deterministic outcomes. We implement a linear annealing strategy for the noise standard deviation.

\xhdr{Data Standardization}
To enhance model stability, we standardize the data before feeding it to the model and remove outlier data points that deviate significantly from the median. Note that scaled and shifted datasets generated by LSNMs or ANMs remain within the same model class.

\xhdr{Numerical Stability and Regularization}
In our implementation, we feed $\scalefunc_{\pi(i)}^\theta(\vect{x}_{< \pi(i)})$ to an exponential function to guarantee positivity. The inherent numerical instability of this function can result in gradient explosion. To mitigate this, we consider an idea similar to \verb|Acrnotm| \citep{kingma2018glow}, where we differentiably scale and shift the activations, using $\tanh$, before passing them through the exponential function. Specifically, denoting the output of previous layers as $h$, we compute $\scalefunc_{\pi(i)}^\theta = \tanh(h / a) \times b$. This ensures that the value lies in the range $[-b, +b]$ with the scaling factor $a$ maintaining sufficiently scaled gradients to prevent them from vanishing. Moreover, we apply an element-wise Sigmoid function on the permutation parameter $\StructureBelief$ to avoid extreme values. We also employ the Adam optimizer with weight decay (AdamW) on the default setting for the training process.

\xhdr{Phase-Changing Training Scheme} We employ a two-phase training approach, alternating between training $\theta$ and $\StructureBelief$. One phase focuses on training $\theta$, where the model learns $\ell_N(\HardPermutation; \theta)$ for $\HardPermutation \in \mathcal{H}_k$, encouraging the model to converge towards an ensemble capable of generating high likelihood values for permutations sourced from $\mathcal{H}_k$. After sufficient training of $\ell_N(\HardPermutation; \theta)$ (signalled by a scheduler in our implementation), we switch to learn $\StructureBelief$, which leads to permutation distributions with higher likelihoods. Note that insufficient training steps for each of the phases can result in error propagation and consequently cause the model to increasingly focus on an erroneous permutation, affecting the overall performance.

\subsection{Evaluation on Real-World and Semi-Synthetic Datasets}

To fairly compare \method\ with other baselines, we use the Sachs and SynTReN datasets to benchmark the results via standard metrics such as SHD and SID. \method\ starts by learning the true ordering and constructs a full tournament graph based on this ordering (where every node is connected to any other node after it in the ordering). This results in a dense graph and to convert the resulting graph into a minimal and sparse one, we apply two pruning techniques: $(i)$ \textbf{CAM pruning}~\citep{buhlmann2014cam}, which employs Lasso sparse regression for pruning. Even though it was originally designed for homoscedastic settings (ANMs), we extend its application to the Sachs and Syntren datasets. $(ii)$ \textbf{PC-KCI}, which leverages conditional independence testing in \citet{zhang2012kernel} to derive a valid causal graph skeleton. Once this skeleton is obtained, we integrate the learned ordering to retrieve the complete graph. For the Sachs dataset, we run the algorithm using $5$ different seeds and report the metrics' average and standard deviation.
\subsection{Baselines} \label{appx:baselines}

To ensure a comprehensive and fair comparison, all baselines are run with \textit{and} without data standardization for the Sachs and Syntren datasets. We run some of the baselines with additional hyperparameter tuning. For \textit{CAM}, we consider both linear and nonlinear regression models. For \textit{DAGuerreo}, we test both SparseMAP and Top-$k$ sparsemax options and linear and nonlinear equation models. Lastly, \textit{VI-DP-DAG} is trained with both Gumbel-Top-$k$ and Gumbel-Sinkhorn alternatives. We also consider a range of epochs, setting a maximum limit of $100$ or $1000$ to assess performance impact.
After testing, we employ all methods with their respective best-performing hyperparameters based on the \cbc\ metric on the Sachs dataset. Once the most effective hyperparameters are determined, we fix these settings for all synthetic data experiments. 

\subsection{The Synthetic Benchmark} \label{appx:synthetic-benchmark}
We devise a comprehensive benchmark suite that provides a flexible and reproducible means of generating diverse datasets based on various graph structures and functional forms. We partition the SCM generation process into four distinct phases:

\begin{enumerate}[leftmargin=*]
\item \textbf{Graph Generation:} Our benchmark generates random graphs of varying sizes with three different types: $(i)$ {Causal paths}, which follow a strict sequence of cause and effect. These graphs are sparse and have unique causal orderings. $(ii)$ {Full graphs (tournaments)} that also possess a unique ordering, where each variable depends on all preceding variables. $(iii)$
{Erdős–Rényi random graphs}, which is based on the graph generation method proposed by \citet{erdHos1960evolution}. These random graphs can have multiple correct orderings.
\item \textbf{Functional Form Generation:}  We use LSNMs (Definition~\ref{def:lsnm}) to model the functional relationships between the parent-children variables:
\begin{equation*}
X_i = \locfunc_i(\vect{X}_{\vect{PA}^\graph_i}) + \scalefunc_i(\vect{X}_{\vect{PA}^\graph_i}) \cdot U_i
\end{equation*}
Our benchmark generates datasets under the following regimes:
\begin{itemize}
\item \textbf{Linear:} Here, both $\locfunc_i$ and $\scalefunc_i$ are random linear combinations of the inputs. To mitigate var-sortability, we normalize the data while simulating the data-generating process along the correct graph ordering \citep{reisach2021beware}.
\item \textbf{Sinusoidal Parametric:} This parametric model is designed to generate datasets satisfying the identifiability criterion specified in Condition~\ref{def:restricted-multi}. In particular, we utilize the concrete conditions discussed in \citet{khemakhem2021causal} for Gaussian exogenous noise and apply the invertible $\sin(x) + x$ function to a linear combination of parents for each covariate to introduce non-linearity into $\locfunc_i$ and enforce positivity in $\scalefunc_i$ using a softplus function.
\item \textbf{Polynomial Parametric:} Similar to the Sinusoidal parametric scheme, we introduce non-linearity into $\locfunc_i$ via a polynomial $x^3 + b$ (where $b$ is randomly selected) and apply a softplus function to ensure $\scalefunc_i$ remains positive. The data is normalized during generation to prevent exponential growth.
\item \textbf{Non-parametric:} Following \citet{zhu2019causal}, we sample functions $\locfunc_i$ and $\scalefunc_i$ from Gaussian processes. To ensure positivity, $\scalefunc_i$ also passes through a softplus.
\end{itemize}
\item 
\textbf{Noise Generation:} To emphasize the potential impacts of model misspecification, we consider both Gaussian and non-Gaussian (specifically, Laplace) noise distributions for our benchmarks.
\item \textbf{Affine/Additive Testing:} We also include simulations where $\scalefunc_i$ is a constant function, enabling a comparison with other baseline models that only work in additive settings.
\end{enumerate}
In all our experiments, we take $1000$ samples from the specified SCMs with different random seeds for graph generation, function creation, and data simulation. A summary of all the datasets incorporated into our benchmark is shown in \autoref{tab:benchmark}.
\begin{table}
\centering
\caption{\small Comparison of baseline methods with \method\ on large synthetic datasets in terms of \cbc, presented as mean $\pm$ standard deviation across samples.}
\label{tab:synthetic-large}
\begin{adjustbox}{width=0.7\linewidth,center}
\small
\begin{tabular}{l*{8}{c}}
\toprule
& \multicolumn{4}{c}{\textbf{Large Datasets}} \\
\cmidrule(lr){2-5}
\textbf{Method} & \multicolumn{2}{c}{\textbf{Parametric}} & \multicolumn{2}{c}{\textbf{Non-parametric}}\\
& \textbf{$d=10$} & \textbf{$d=25$} & \textbf{$d=10$} & \textbf{$d=25$} \\
\midrule
\method~(Gumble-Top-$k$)   & $\mathbf{0.16 \pm 0.08}$ & $\mathbf{0.24 \pm 0.03}$ & $\mathbf{0.2 \pm 0.06}$ & $0.54 \pm 0.05$ \\
\midrule
CAM        & $0.85 \pm 0.12$ & $0.9 \pm 0.07$ &  $0.27 \pm 0.23$ & $0.28 \pm 0.18$ \\
VI-DP-DAG   & $0.64 \pm 0.22$ & $0.59 \pm 0.11$ & $0.46 \pm 0.23$ & $0.44 \pm 0.12$ \\
DAGuerreo        & $0.88 \pm 0.07$ & $0.88 \pm 0.04$ & $0.3 \pm 0.07$ & $0.38 \pm 0.06$ \\
SCORE      &  $0.9 \pm 0.1$ & $0.96 \pm 0.03$ &  $0.2 \pm 0.09$ & $\mathbf{0.17 \pm 0.06}$ \\
VarSort     & $0.86 \pm 0.08$ & $0.84 \pm 0.11$ & $0.49 \pm 0.11$ &  $0.55 \pm 0.07$ \\
biLSNM     & $0.62 \pm 0.09$ & $0.52 \pm 0.04$  & $0.65 \pm 0.12$ & $0.54 \pm 0.03$ \\
\bottomrule
\end{tabular}%
\end{adjustbox}
\end{table}

\begin{table}
\centering
\caption{\small Summary of the Synthetic Benchmark Datasets.}
\label{tab:benchmark}
\renewcommand{\arraystretch}{1.5}
\begin{tabularx}{\textwidth}{|>{\centering\arraybackslash}X|>{\centering\arraybackslash}X|>{\centering\arraybackslash}X|>{\centering\arraybackslash}X|>{\centering\arraybackslash}X|>{\centering\arraybackslash}X|}
\hline
\textbf{Affine or Additive} & \textbf{Functional Form} & \textbf{Graph Size} & \textbf{Graph Type} & \textbf{Noise} & \textbf{\# of Simulations} \\
\hline
Affine & Nonparametric & $3 \le d \le 6$ & Erdős–Rényi & Normal & 20 \\
\hline
Affine & Nonparametric & $3 \le d \le 6$ & Tournament & Normal & 20 \\
\hline
Affine & Nonparametric & $3 \le d \le 6$ & Path & Normal & 20 \\
\hline
Additive & Nonparametric & $3 \le d \le 6$ & Erdős–Rényi & Normal & 20 \\
\hline
Additive & Nonparametric & $3 \le d \le 6$ & Tournament & Normal & 20 \\
\hline
Additive & Nonparametric & $3 \le d \le 6$ & Path & Normal & 20 \\
\hline
Affine & Linear & $3 \le d \le 6$ & Erdős–Rényi & Laplace & 20 \\
\hline
Affine & Linear & $3 \le d \le 6$ & Tournament & Laplace & 20 \\
\hline
Affine & Linear & $3 \le d \le 6$ & Path & Laplace & 20 \\
\hline
Additive & Linear & $3 \le d \le 6$ & Erdős–Rényi & Laplace & 20 \\
\hline
Additive & Linear & $3 \le d \le 6$ & Tournament & Laplace & 20 \\
\hline
Additive & Linear & $3 \le d \le 6$ & Path & Laplace & 20 \\
\hline
Affine & Nonparametric & $d \in \{10, 25\}$ & Erdős–Rényi & Normal & 10 \\
\hline
Affine & Sinusoidal & $d \in \{10, 25\}$ & Erdős–Rényi & Normal & 10 \\
\hline
Affine & Polynomial & $d \in \{10, 25\}$ & Erdős–Rényi & Normal & 10 \\
\hline
Affine & Sinusoidal & $3 \le d \le 6$ & Erdős–Rényi & Normal & 20 \\
\hline
Affine & Sinusoidal & $3 \le d \le 6$ & Tournament & Normal & 20 \\
\hline
Affine & Sinusoidal & $3 \le d \le 6$ & Path & Normal & 20 \\
\hline
Affine & Polynomial & $3 \le d \le 6$ & Erdős–Rényi & Normal & 20 \\
\hline
Affine & Polynomial & $3 \le d \le 6$ & Tournament & Normal & 20 \\
\hline
Affine & Polynomial & $3 \le d \le 6$ & Path & Normal & 20 \\
\hline
Affine & Linear & $3 \le d \le 6$ & Erdős–Rényi & Normal & 20 \\
\hline
Affine & Linear & $3 \le d \le 6$ & Tournament & Normal & 20 \\
\hline
Affine & Linear & $3 \le d \le 6$ & Path & Normal & 20 \\
\hline
Additive & Sinusoidal & $3 \le d \le 6$ & Erdős–Rényi & Normal & 20 \\
\hline
Additive & Sinusoidal & $3 \le d \le 6$ & Tournament & Normal & 20 \\
\hline
Additive & Sinusoidal & $3 \le d \le 6$ & Path & Normal & 20 \\
\hline
Additive & Polynomial & $3 \le d \le 6$ & Erdős–Rényi & Normal & 20 \\
\hline
Additive & Polynomial & $3 \le d \le 6$ & Tournament & Normal & 20 \\
\hline
Additive & Polynomial & $3 \le d \le 6$ & Path & Normal & 20 \\
\hline
Additive & Linear & $3 \le d \le 6$ & Erdős–Rényi & Normal & 20 \\
\hline
Additive & Linear & $3 \le d \le 6$ & Tournament & Normal & 20 \\
\hline
Additive & Linear & $3 \le d \le 6$ & Path & Normal & 20 \\
\hline
\end{tabularx}
\end{table}

The results of our synthetic benchmark for small graphs are provided in \autoref{tab:synthetic}. To evaluate the scalability of our model, we also consider large, random graphs with $d \in \{10, 25\}$, as presented in \autoref{tab:synthetic-large}. The results reveal that \method\ consistently outperforms others when handling data in parametric function forms. In non-parametric scenarios, it performs equally well as the top baselines for $d=10$ but shows less efficacy for $d=25$. We hypothesize that increasing the number of epochs could improve this performance. However, these findings also suggest certain limitations of \method\ in scaling.

%% file: appendices/interventions.tex
\section{Interventional Distributions} \label{appx:interventions}
We consider synthetically generated datasets to investigate \method's capability in estimating downstream interventional distributions. Our practical observations suggest that uncovering a valid causal ordering can be sufficient for estimating average causal effects.

\xhdr{Dataset} We mainly follow similar LSNM data generation as described in our synthetic benchmark in Appendix~\ref{appx:synthetic-benchmark}. In particular, we consider two types of sinusoidal functions for the location, $\locfunc(x) = x + \sin(x)$ and $\locfunc(x) = \sin(ax) + \sin(bx)$, where the latter is non-invertible and set the scale function as $\scalefunc(x) = \alpha - \beta e^{-\frac{|x|}{2}}$ ($\alpha > \beta$). Moreover, we examine two types of causal graphs, tournaments and causal paths, to capture the various dependencies that can occur when computing interventions. Finally, we choose the base distribution as standard Gaussian.

\begin{figure}[h!]
\captionsetup{font=footnotesize,labelfont=footnotesize}
    \centering
    \includegraphics[width=\textwidth]{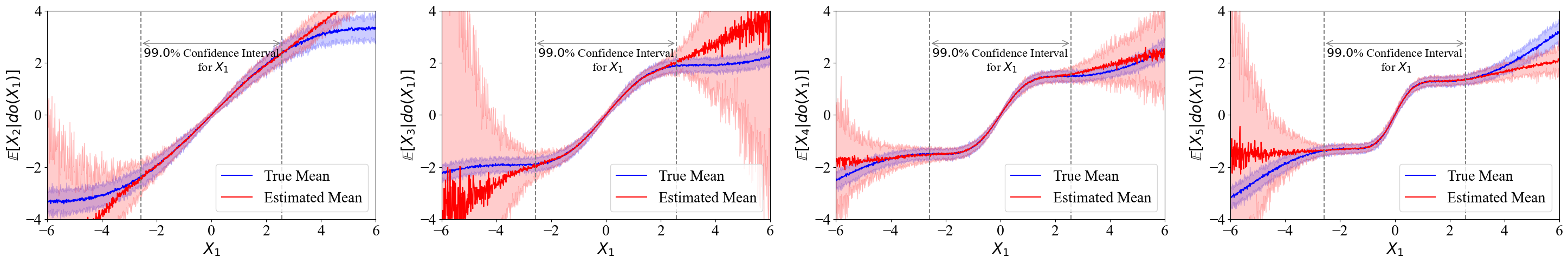}
    \caption{\small Estimating the interventional expected value $\E[X_2, \ldots, X_5 | do(X_1)]$ using \method\ in a {tournaments} causal graph from $X_1$ to $X_5$. The estimated value matches the true expectation in the $99\%$ confidence interval of the observational data.}
    \label{fig:intervention_sup_grid}
\end{figure}

\begin{figure}[h!]
\captionsetup{font=footnotesize,labelfont=footnotesize}
    \centering
\includegraphics[width=0.7\textwidth]{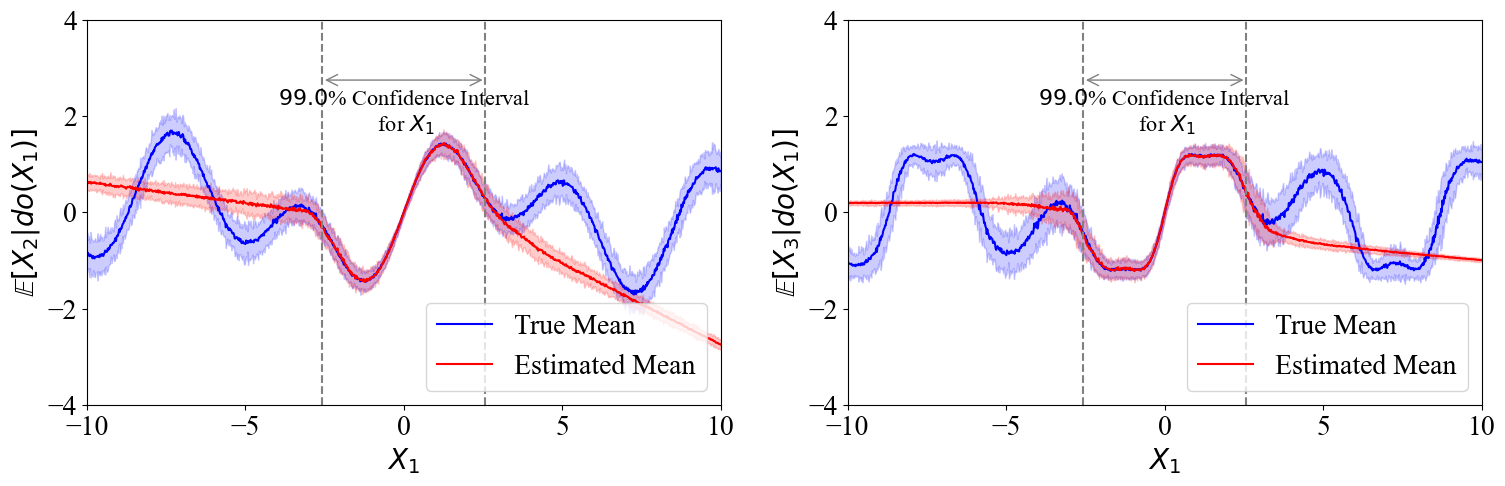}
    \caption{\small Estimation of interventional expected values in a causal path with $X_1$ and $X_3$ being the first and last nodes.}
    \label{fig:intervention_sup_grid2}
\end{figure}

\xhdr{Estimating Interventional Distributions} To independently explore the properties of intervention estimation, we assume the true causal ordering $\pi$ is either known or identified by running \method. Given $\pi$, we train \method\ to minimize the negative-log-likelihood of observational samples generated from the aforementioned dataset. For ease of notation, we assume the correct causal ordering as ${1 < 2 < \cdots < d}$. Our goal is to estimate the expected causal effect with hard interventions $\E(\vect{X}|do(X_1)=y)$. Given a trained ANF model $\vect{X}=\vect{T}^{\theta}(\vect{U})$, we can sample from the interventional distribution using the procedure described in Algorithm~\ref{algo:intervention}, a similar approach to \citet{khemakhem2021causal}. Specifically, we use $50$ independently drawn samples from the interventional distribution to estimate its mean and the $99\%$ confidence interval.
\begin{algorithm}[ht]
  \SetKwInput{KwData}{Input}
  \SetKwInput{KwResult}{Output}
  \SetAlgoLined
  
  \KwData{The value of $y$ for $do(X_1)=y$}
  \KwResult{A single sample drawn from $P(\vect{X}|do(X_1)=y)$}
  
  \BlankLine
    $\vect{U} \sim P_{\vect{U}}$ \hfill \#{sample the noises}
    
  $U_1 \leftarrow T_1^{-1}(y)$ \hfill \#{ replace $U_1$ with the associated noise which results in $X_1=y$}
  
  \Return{$\vect{T}(\mathbf{U})$ \hfill \#{ a sample from the hard interventional distribution}}
  
  \caption{Generating samples from the hard interventional distribution}
    \label{algo:intervention}
\end{algorithm}~\vspace{-1em}
\xhdr{Results}
\autoref{fig:intervention} and \autoref{fig:intervention_sup_grid} illustrate the estimated interventional expected values of \method\ in the tournament causal graph using $4000$ observational samples. We observe that the trained model can accurately estimate the interventional distributions in the $99\%$ confidence interval of the observational data. Similarly, \autoref{fig:intervention_sup_grid2} demonstrate the capability of \method\ in estimating interventional distributions in a causal path graph with three nodes. Note that even though a large enough model can accurately estimate the average causal effects, the predictions significantly deviate from the ground-truth value outside of the observational regime.
